\documentclass[lettersize, journal]{IEEEtran}
\usepackage{multirow}
\usepackage{color}
\usepackage{hyperref}
\usepackage[normalem]{ulem}
\usepackage{balance}
\usepackage{cases}
\usepackage{array}
\usepackage{bbding}
\usepackage{amssymb}
\usepackage{graphicx}
\usepackage{booktabs}
\usepackage{diagbox}
\usepackage{caption}
\usepackage{subcaption}
\usepackage{graphicx}

\usepackage{orcidlink}
\ifCLASSINFOpdf
\else
\fi
\begin{document}

\title{Image-based Freeform Handwriting Authentication with Energy-oriented Self-Supervised Learning}
% Towards Robust and Fast Freeform Handwriting Authentication with Energy-oriented Two-branch Contrastive Self-Supervised Learning

\author{Jingyao~Wang,
        Luntian~Mou,~\IEEEmembership{Senior Member,~IEEE,}
        Changwen~Zheng,
        and Wen~Gao,~\IEEEmembership{Fellow, IEEE} % <-this % stops a space
\thanks{
    This work was supported in part by the National Natural Science Foundation of China under Grant 61672068, and in part by the Beijing Natural Science Foundation under grant 4242017 (Corresponding authors: Luntian Mou and Changwen Zheng).
    
    J. Wang and C. Zheng are with the University of Chinese Academy of Sciences, Beijing 100049, China. They are also with the National Key Laboratory of Space Integrated Information System, Institute of Software Chinese Academy of Sciences, Beijing 100086, China. E-mail: wangjingyao22@mails.ucas.ac.cn, changwen@iscas.ac.cn. 

    L. Mou is with Beijing Institute of Artiicial Intelligence, Faculty of Information, Beijing University of Technology, Beijing 100124, China. He is also with Beijing Key Laboratory of Multimedia and Intelligent Software Technology, Beijing 100124, China. E-mail: ltmou@pku.edu.cn.
    
    W. Gao is with the School of Electronics Engineering and Computer Science, Peking University, Beijing 100871, China. E-mail: wgao@pku.edu.cn.

    }      
}

% % \markboth{IEEE Transactions on Circuits and Systems for Video Technology}%
% \markboth{IEEE Transactions on Multimedia}%
% {Shell \MakeLowercase{\textit{et al.}}: Bare Demo of IEEEtran.cls for IEEE Journals}

\markboth{IEEE TRANSACTIONS ON MULTIMEDIA}{Wang \MakeLowercase{\textit{et al.}}: Image-based Freeform Handwriting Authentication with Energy-oriented Self-Supervised Learning}


\maketitle

\begin{abstract}
Freeform handwriting authentication verifies a person’s identity from their writing style and habits in messy handwriting data. This technique has gained widespread attention in recent years as a valuable tool for various fields, e.g., fraud prevention and cultural heritage protection. However, it still remains a challenging task in reality due to three reasons: (i) severe damage, (ii) complex high-dimensional features, and (iii) lack of supervision. To address these issues, we propose SherlockNet, an energy-oriented two-branch contrastive self-supervised learning framework for robust and fast freeform handwriting authentication. It consists of four stages: (i) pre-processing: converting manuscripts into energy distributions using a novel plug-and-play energy-oriented operator to eliminate the influence of noise; (ii) generalized pre-training: learning general representation through two-branch momentum-based adaptive contrastive learning with the energy distributions, which handles the high-dimensional features and spatial dependencies of handwriting; (iii) personalized fine-tuning: calibrating the learned knowledge using a small amount of labeled data from downstream tasks; and (iv) practical application: identifying individual handwriting from scrambled, missing, or forged data efficiently and conveniently. Considering the practicality, we construct EN-HA, a novel dataset that simulates data forgery and severe damage in real applications. Finally, we conduct extensive experiments on six benchmark datasets including our EN-HA, and the results prove the robustness and efficiency of SherlockNet. We will release our code and dataset at \href{https://github.com/WangJingyao07/SherlockNet}{https://github.com/WangJingyao07/SherlockNet}.
\end{abstract}


\begin{figure}
  \includegraphics[width=0.5\textwidth]{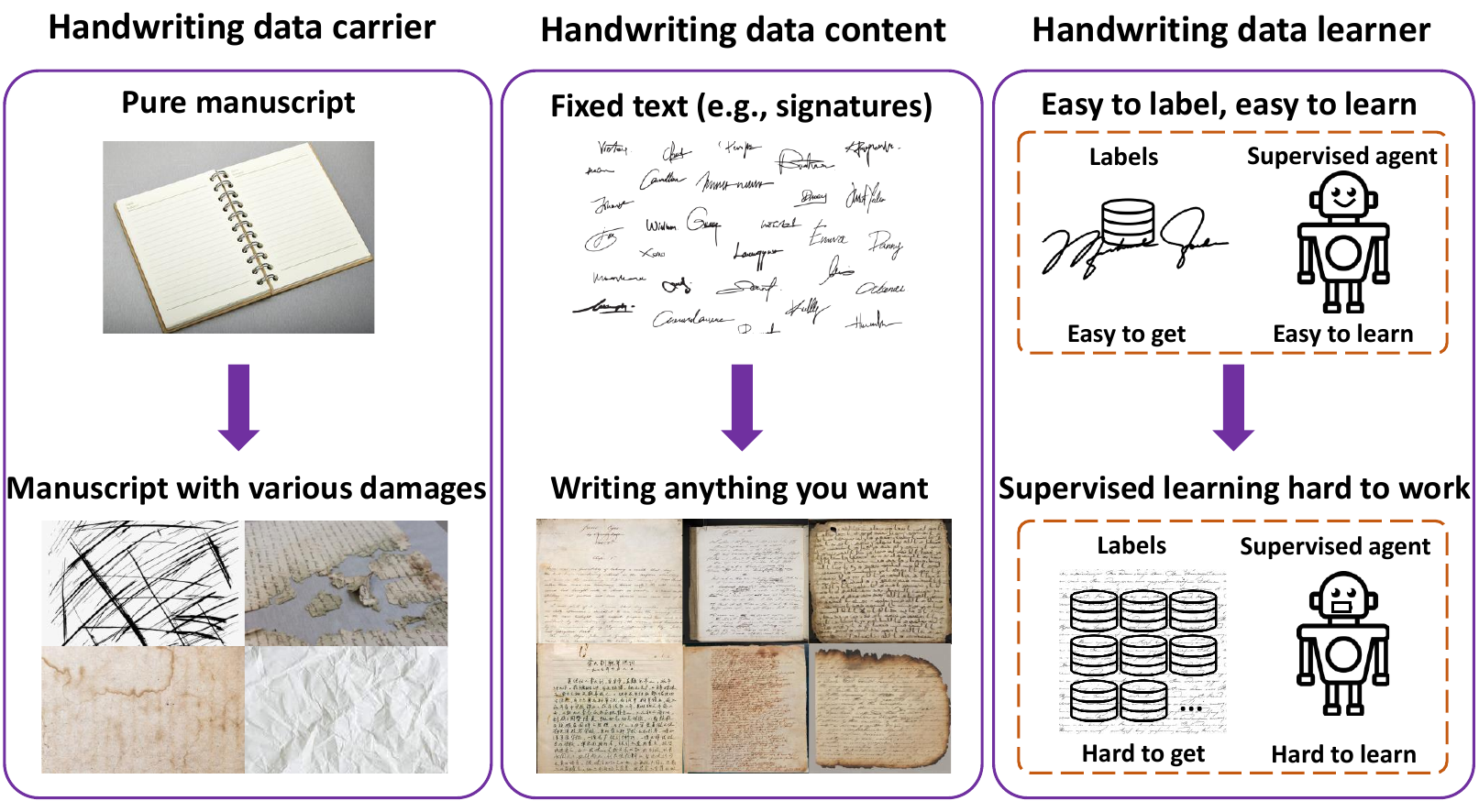}
    \caption{Handwriting authentication vs. freeform handwriting authentication. Compared with previous handwriting authentication, the challenging FHA requires a model: (i) not restricting data quality; (ii) not constraining handwriting content; and (iii) not relying on supervision information.}
    \label{fig:intro_1}
    \vspace{-0.1in}
\end{figure}

\section{Introduction}
\label{sec:1}
\IEEEPARstart{H}{andwriting} refers to the act of writing using a specific instrument (e.g., a pen or pencil). Since each person's handwriting is unique, it is considered an important characteristic that is used in applications of various fields such as identity verification, e-security, and e-health \cite{li2022fast, yun2021instance, santangelo2016comprehensive}. For instance, doctors utilize changes in handwriting as a diagnostic sign for Neurodegenerative Diseases (NDs) \cite{de2019handwriting}, while forensic experts use handwriting clues to identify suspects \cite{morris2020forensic}. Therefore, handwriting authentication \cite{zhuang2023optimizing,al2019new,begum2021user,wang2023cssl} emerged as a result and became an interdisciplinary topic, aiming to verify or identify individuals using the unique features and patterns of handwriting. However, existing methods use a supervised manner for handwriting authentication that requires annotated handwriting data, which costs intensive \cite{hirata2015matching, sharipov2019analysis, faundez2020handwriting}.

Imagine creating archives from written materials such as books, papers, or ancient bamboo slips. These archives contain large amounts of unlabeled handwriting data uploaded from various individuals in the real world, where the manual annotation process is complex and lengthy. Moreover, the real-world handwriting data is prone to noise and corruption \cite{gonwirat2022deblurgan,li2022fast,yun2021instance}, so many supervised methods that work on labeled and clean data may fail \cite{dong2022seismic}. To address these challenges, self-supervised learning (SSL) has become popular in recent years \cite{jaiswal2020survey, misra2020self}. This paradigm aims to learn general representations without supervision and adapt well to downstream tasks.

Several studies have delved into self-supervised learning to acquire representations of handwriting features, which can overcome the scarcity of supervision \cite{lastilla2022self, chattopadhyay2022surds, gidaris2020learning}. Gidaris et al. \cite{gidaris2020learning} encoded discrete visual concepts with spatially dense image descriptions. Lastilla et al. \cite{lastilla2022self} used unlabeled pure manuscripts to learn texture features. Chattopadhyay et al. \cite{chattopadhyay2022surds} leveraged SSL and metric learning for writer-independent OSV in a two-stage deep learning framework.
However, in cases where the data is damaged or encounters significant disturbances, obtaining representations of the original data can result in significant ambiguities \cite{fang2022corrupted, xu2020noisy}. 
This issue is particularly challenging in handwriting authentication, as it involves a vast number of damaged and fake samples in the real world, while the handwriting features are also complex.

In this work, we focus on a more challenging handwriting authentication topic: freeform handwriting authentication (see Figure \ref{fig:intro_1}). It requires the model to perform accurate identity verification via freeform handwriting rather than just fixed content, facing huge challenges. Firstly, the handwriting data in real life may suffer from various damages \cite{li2023improving,zhao2023handwriting}, such as scratches, stains, and folds from improper paper storage. Secondly, freeform handwriting means that the features become more complex and diverse. Unconstrained handwriting content also means that there may be multiple similar characters in the handwriting data, but with different strokes, colors, etc., which also makes handwriting authentication more difficult. In addition, due to the richness of the handwriting content, the annotation will face greater cost pressure \cite{vijayanarasimhan2009s, wu2017diverse}, making the supervised paradigm more unrealistic. Therefore, a superior model needs to achieve robust and fast freeform handwriting authentication under the conditions of (i) not restricting data quality; (ii) not constraining handwriting content; and (iii) not relying on supervision information.

To address these challenges, we propose SherlockNet, a novel energy-oriented two-branch contrastive self-supervised learning framework that focuses on the robustness, efficiency, and practicality in freeform handwriting authentication. It consists of four stages: pre-processing, generalized pre-training, personalized fine-tuning, and practical application. Specifically, in the pre-processing stage, we propose a plug-and-play energy operator that converts handwriting manuscripts into a series of energy distributions, effectively filtering out noise and preserving stroke information. The energy value indicates the probability that it belongs to handwriting features instead of noise. In the generalized pre-training stage, we propose a two-branch momentum-based adaptive contrastive learning framework to learn general representations from energy distributions, quickly extracting high-dimensional features of handwriting and mining spatial dependencies. In the personalized fine-tuning stage and practical application stage, we design convenient interfaces to make the users deploy the proposed SherlockNet in any downstream scenario, which only relies on a few samples with a few steps. %The brief description of our proposed SherlockNet is depicted in Figure \ref{fig:teaser}.

Our contributions can be summarized as follows:
\begin{itemize}
\item We explore a challenging topic: freeform handwriting authentication, which faces three key issues in real-world applications: (i) severe damage, (ii) complex high-dimensional features, and (iii) lack of supervision.
\item We propose SherlockNet, an energy-oriented two-branch contrastive self-supervised learning framework for robust and fast freeform handwriting authentication.
\item We present EN-HA, a freeform handwriting authentication dataset that mimics real-world scenarios, such as data forgery and severe damage in handwriting data. 
\item Extensive experiments are conducted on six benchmark datasets, and the results demonstrate the superior robustness and efficiency of the proposed SherlockNet.
\end{itemize}

%% ----------------Figure 2-----------------------
\begin{figure*}
    \centering
    \includegraphics[width=0.9\textwidth]{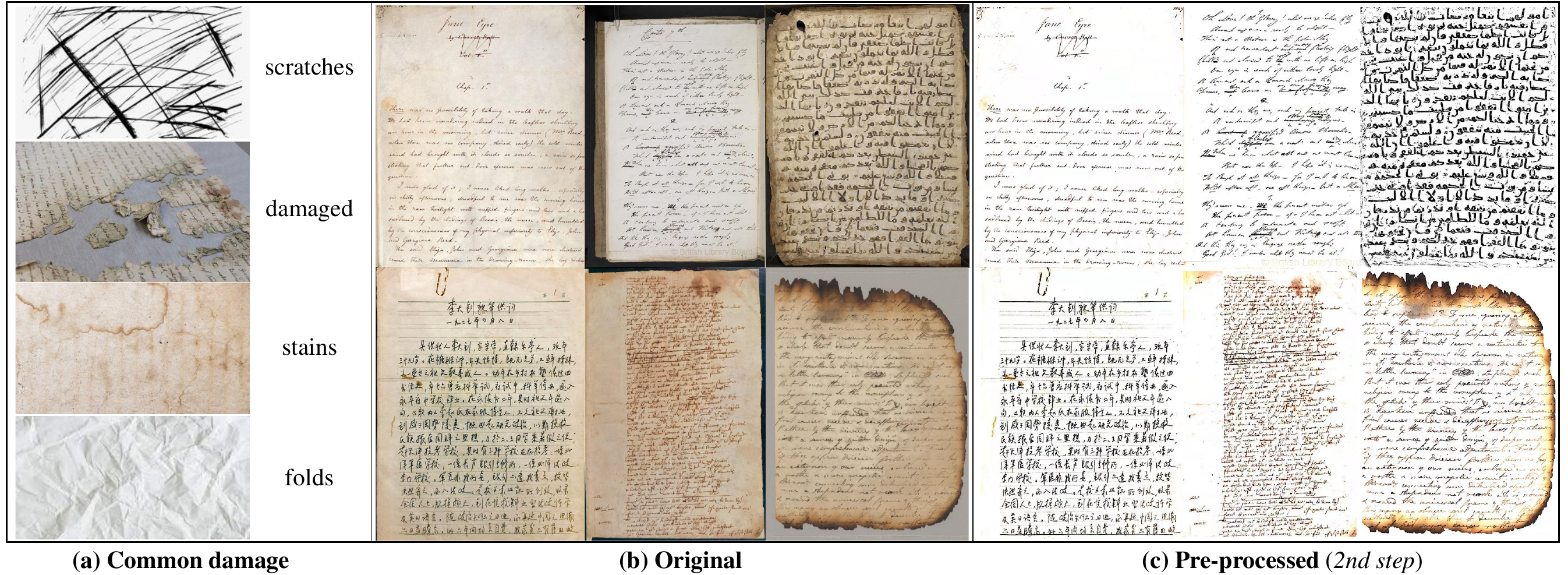}
    \caption{Handwriting defects and pre-processing results. (a) the common defects and damages of the collected manuscripts; (b) samples of the collected handwriting data in the real world; (c) samples after pre-processing with 2 steps.}
\label{fig:defect}
\end{figure*}

%% -----------------------RELATED WORK--------------------------------
\section{Related Work}
\label{sec:2}

\subsection{Handwriting Authentication}
\label{sec:2.1}
Unlike physiological characteristics, handwriting is a behavioral characteristic, where no two individuals with identical handwriting, and one individual cannot produce another's handwriting exactly \cite{rehman2019writer}. Handwriting has gained increasing attention and found widespread applications in various fields such as forensic identification \cite{morris2020forensic}, medical diagnosis \cite{de2019handwriting}, information security \cite{faundez2020handwriting}, and document forgery detection \cite{adak2019empirical}. However, due to its complexity and large variability, handwriting authentication remains a challenging task \cite{al2019new,begum2021user}.

To cope with the growing demand for handling big data, traditional manual inspection of handwriting is inefficient \cite{anikin2022framework, gupta2020improved}. As a result, research has shifted towards machine-based methods \cite{li2021avn, liu2020parameter, zhang2020srd} to bridge the gap between current needs and actual applications. Davis et al. \cite{davis2020text} developed a generator network trained with GAN and autoencoder techniques to learn handwriting style. Zhang et al. \cite{zhang2016end} proposed an end-to-end framework for online text-independent writer identification that leverages a recurrent neural network (RNN). Mohammad et al. \cite{al2019new} proposed a new Arabic online/offline handwriting dataset and using an SVM-based method for writer authentication. Begum et al. \cite{begum2021user} concentrated on real-time and context-free handwriting data analysis for robust user authentication systems using digital pen-tablet sensor data. Although these methods have been successful in enhancing the efficiency of handwriting authentication, they rely on a large amount of labeled and pure data, while having high requirements on the writing content, for example, it can only be signature. Unfortunately, in reality, it is difficult to meet the data requirements of these methods (see Figures \ref{fig:intro_1} and \ref{fig:defect}).

In this study, considering practicality in the real world, we focus on and explore a more challenging topic for the first time, i.e., freeform handwriting authentication. It is more challenging in that: (i) it does not rely on supervised information; (ii) it does not restrict the handwriting content; and (iii) it does not require pure data, which makes this work novel and practical, and can be applied to real-world scenarios.

\subsection{Self-supervised Learning}
\label{sec:2.2}

Self-supervised learning (SSL) plays a crucial role in advancing technology during the big data era, offering robust representation by learning knowledge from pretext tasks without annotations. Notably, SSL methods can be broadly divided into two types: contrastive learning and generative learning.

Contrastive learning performs learning by applying various data augmentations and encouraging the augmented views of the same sample closer while pushing apart views from other samples, and then using the learned knowledge to solve new tasks \cite{shurrab2022self}. This technique can significantly enhance performance in various visual tasks \cite{LMLC, mou2021isotropic}, and its typical algorithms include SimCLR \cite{simclr}, MoCo \cite{moco}, BYOL \cite{byol} etc. For handwriting data, Lin et al. \cite{lin2022cclsl} proposed using contrastive learning to learn semantic-invariant features between printed and handwriting characters. Viana et al. \cite{viana2022contrastive} explained how deep contrastive learning can lead to an improved dissimilarity space for learning representations from handwriting signatures. Zhang et al. \cite{zhang2022cmt} extended Moco \cite{moco} to show the significant advantages of self-supervision in dealing with complex handwriting features. Although the existing methods have demonstrated the ability to extract complex features, they rely on a large amount of pure handwriting data with consistent content, making it difficult to apply well and achieve free-form handwriting authentication in the real world.

Generative learning involves the reconstruction of inputs from original or damaged inputs, and the creation of representation distributions at a point-wise level, such as pixels in images and nodes in graphs. Classic methods include auto-regressive (AR) models \cite{matero2020autoregressive, you2018graphrnn}, flow-based models \cite{dohi2021flow, kingma2018glow}, auto-encoding (AE) models \cite{kipf2016variational, sabokrou2019self}, and hybrid generative models \cite{dai2019transformer, khajenezhad2020masked}. For handwriting data, considering the properties of this paradigm, generative learning can work in terms of data augmentation for auxiliary training when data is insufficient \cite{wu2021self,eckart2021self,von2021self}. He et al. \cite{he2022masked} introduced masked autoencoders (MAE) as a scalable self-supervised learning approach for computer vision aimed at reconstructing missing patches in images. However, although generative learning can mitigate the negative effects of lacking labels, it is not suitable for handwriting authentication whose primary focus is learning complex features without supervision.

%% ------Preliminary----------
\section{Preliminary}
\label{sec:3}
In this section, we first provide the preliminary of contrastive self-supervised learning. Next, we introduce the formulation of our SherlockNet. We adhere to the standard contrastive self-supervised learning pipeline, which involves self-supervised pre-training followed by task-specific fine-tuning. 

\subsection{Pre-training}
\label{sec:3.1}
Given a dataset $ \mathcal{D}= \{ x_i  \}_{i=1}^N$ and a augmentation distribution $\mathcal{A}=\{ a^j(x_i)  \}_{i=1,j=1}^{i=N,j=2}$, the contrastive SSL model is denoted as $f_\theta=h\circ  g$ with a feature extractor $g(\cdot)$ and a projection head $h(\cdot)$. The feature extractor, $g(\cdot)$, comprises a foundational neural network structure, while the projection head, $h(\cdot)$, can take the form of either a multi-layer perceptron (MLP) or a single-layer weight matrix. Then, the learning objective of contrastive learning can be expressed as:
\begin{equation}\label{eq:contrastive}
\begin{array}{l}
    \min_{f_{\theta}}\mathcal{L}_{\mathrm{contrastive}}(f_{\theta};\mathcal{D},\mathcal{A}) :=\mathbb{E}_{\mathcal{B}_{x} \sim \mathcal{D}}\left [ \ell(f_{\theta};\mathcal{B}_{x}) \right ] 
\end{array}
\end{equation}
where $x=\{x_i\}_{i=1}^{B}\in\mathcal{B}_{x}\ $ stands for the data within a mini-batch $\mathcal{B}_{x}\in \mathcal{D}$. The function $\ell(\cdot)$ is the InfoNCE loss:
\begin{equation}\label{eq:infonce}
    \ell(f_{\theta};\mathcal{B}_{x}) = -\sum_{z^j_{x_i}\in\mathcal{Z}_{x}}\log\frac{\exp(\cos(z^j_{x_i}, z^{3-j}_{x_i})/\tau)}{\sum_{\mathcal{Z}_{x} \setminus \{z^j_{x_i}\}}\exp(\cos(z^j_{x_i}, z_{x})/\tau)}
\end{equation}
where $z^j_{x_i}=h(g(a^j(x_i)))=f_\theta(a^j(x_i))\in \mathcal{Z}_{x}$ represents the embedding of the augmented data $a^j(x_i)$, which is also the embedding of the anchor, and $z^{3-j}_{x_i}$ is the embedding of the positive sample linked to the anchor. $\cos(u,v) = u^\top v/ \left \| u \right \|   \left \| v \right \|$, $\tau$ serves as a hyperparameter for temperature scaling. This objective (Eq.\ref{eq:contrastive} and Eq.\ref{eq:infonce}) aims to learn representation by increasing the similarity between $z^j_{x_i}$ and $z^{3-j}_{x_i}$, while simultaneously reducing the similarities between $z^j_{x_i}$ and the embeddings of negative samples within $\mathcal{Z}_{x} \setminus \{z^j_{x_i}, z^{3-j}_{x_i}\}$. 

\subsection{Fine-tuning}
\label{sec:3.2}
Once pre-training is completed, the projection head $h(\cdot)$ is discontinued, and the adjustable weights within the feature extractor $g(\cdot)$ become fixed. A coefficient matrix $\mathcal{M} \subseteq \mathbb{R}^{n\times d}$ is then trained utilizing the training set $\mathcal{D}_{tr}^{ds}$ specific to the downstream task, where $n$ and $d$ represent the quantity and embedding dimensions of the samples, respectively. Note that the training set $\mathcal{D}_{tr}^{ds}$ only contains a few samples but each sample is with the corresponding labels. Then, the learning objective of this stage can be articulated as:
\begin{equation}
\label{eq:downstream}
\begin{array}{l}
\arg \min_{\mathcal{M}} \mathcal{L}_{\mathrm{ds}}(\mathcal{M};f_\theta, \mathcal{D}_{tr}^{ds})\\[8pt]
s.t. \quad \mathcal{L}_{\mathrm{ds}}(\mathcal{M};f_\theta, \mathcal{D}_{tr}^{ds})=-\sum_{{x_i^{ds}}\in x}\log P(y^{ds}_{i}; \mathcal{M} f_{\theta}(x^{ds}_{i}))
\end{array}
\end{equation}
where $x_i^{ds}$ and $y_i^{ds}$ are the training sample and the corresponding label. This fine-tuning process consists of only a few optimization steps.

\subsection{Formulation of SherlockNet}
\label{sec:3.3}
We combine the unlabeled training data from all writers denoted as $I=\left \{ \mathcal{I} _{1},...,\mathcal{I} _{N} \right \}\in \mathbb{R}^{H\times W\times C}$, where $N$ is the number of the writers, $H\times W$ denotes the resolution of the images, and $C=2$ represents the number of the branches. The data is pre-processed with an energy operator $S(\mathcal{I}_i)$, and the output denoised data with energy distributions are expressed as $X=\left \{ x_{1},...,x_{N} \right \} \in \mathbb{R}^{H\times W\times C}$. The fine-tuning samples and the corresponding label are represented as $I^{ds}$ and $Y^{ds}$. Turning to the model structure, our proposed SherlockNet is defined by $f_{\theta}$ which consists of two branches, i.e., a contrastive learning branch and a momentum branch. The pre-trained feature extractor is denoted as $g(\cdot)$, which generates the representations $Z_{x}=g(x)$ for a mini-batch of data $x=\left \{ x_{i} \right \}_{i=1}^n$. The contrastive learning branch comprises a pre-trained feature extractor $g(\cdot)$, a patch head $P^{pat}(\cdot)$, a projection head $P^{pro}(\cdot)$, and a prediction head $P^{pre}(\cdot)$. The momentum branch is composed of the same extractor $g(\cdot)$, a patch head $K^{pat}(\cdot)$, and a projection head $K^{pro}(\cdot)$. During the task-specific fine-tuning phase, the patch head, projection head, and prediction head are discarded, while the trainable weights in feature extractor $f_\theta$ are frozen, only a coefficient matrix $\mathcal{W}$ is trained as mentioned in Subsection \ref{sec:3.2}.

%% ------Methodology---------
\section{Methodology}
\label{sec:4}
\begin{figure*}
  \includegraphics[width=\textwidth]{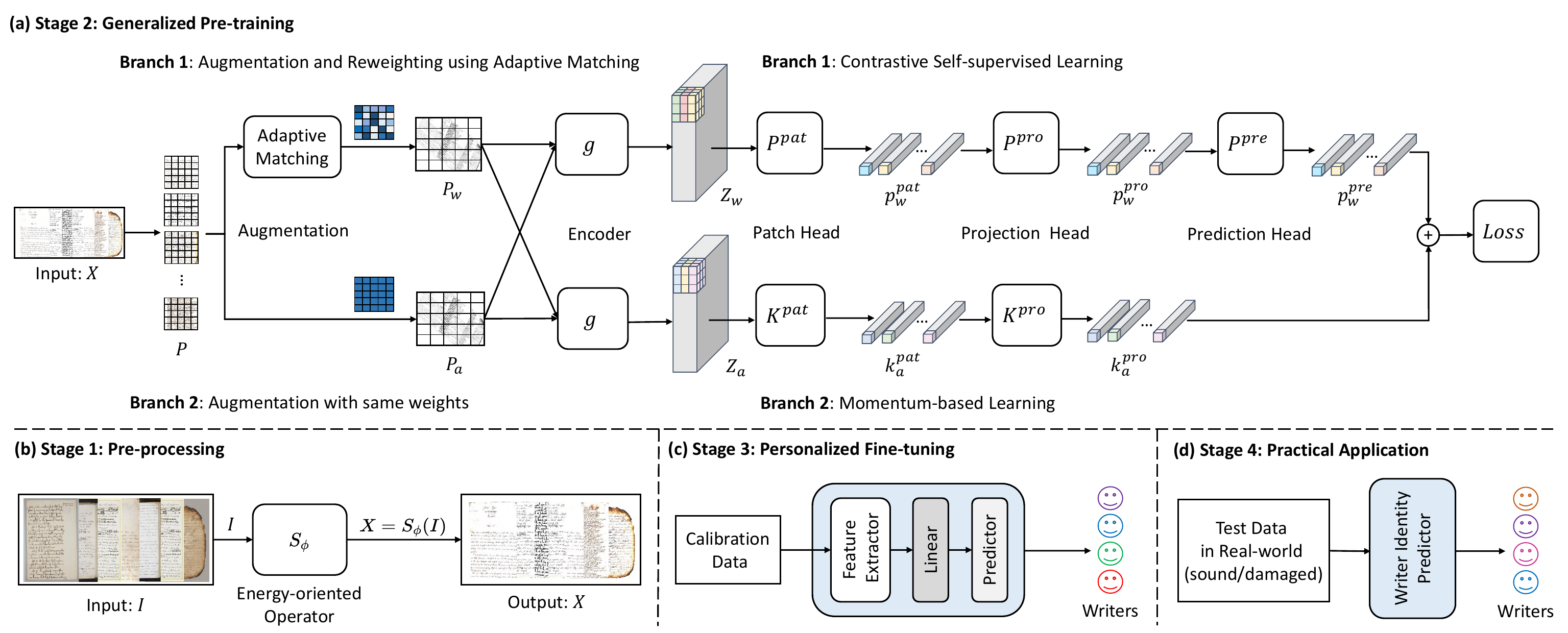}
    \caption{The framework of the proposed SherlockNet with four stages, i.e., pre-processing stage (b), generalized pre-training stage (a), personalized fine-tuning stage (c), and practical application stage (d).}
    \label{fig:SherlockNet}
\end{figure*}

In this section, we introduce SherlockNet, the energy-oriented two-branch contrastive self-supervised learning framework we proposed for robust and fast freeform handwriting authentication. Firstly, we provide the overview of SherlockNet in Subsection \ref{sec:4.1}, which consists of four stages including pre-processing, generalized pre-training, personalized fine-tuning, and practical application. Next, we provide the details of these four stages in Subsections \ref{sec:4.2}-\ref{sec:4.5}, respectively. Figure \ref{fig:SherlockNet} shows the framework of our SherlockNet.

\subsection{Overview}
\label{sec:4.1}
SherlockNet consists of four stages: pre-processing, generalized pre-training, personalized fine-tuning, and practical application. In the pre-processing stage, we use a plug-and-play energy-oriented operator $S_{\phi}$ to convert handwriting manuscripts into energy distributions, filtering out noise based on the calculated energy distributions. The unlabeled handwriting data $I$ from all writers undergo processing to mitigate any negative impacts caused by paper damage, stains, or other defects, obtaining $X$. In the pre-training stage, the pre-processed images $X$ are augmented, reweighted, and compared with each other to learn the general representation by $f_{\theta}$. In the personalized fine-tuning stage, only a small amount of labeled data $\mathcal{D}^{ds}_{tr}$ from a specific downstream task is used to calibrate the personal handwriting predictor from the pre-trained generalized model $f_{\theta}$. In the practical application stage, we evaluate the effect of the model on the test set $\mathcal{D}^{ds}_{te}$ of the downstream task to ensure it can be used efficiently and conveniently to recognize the writer's identities efficiently.

%% ----------------Figure 3-----------------------

%% ----------------Figure 4-----------------------
\begin{figure}
  \includegraphics[width=0.5\textwidth]{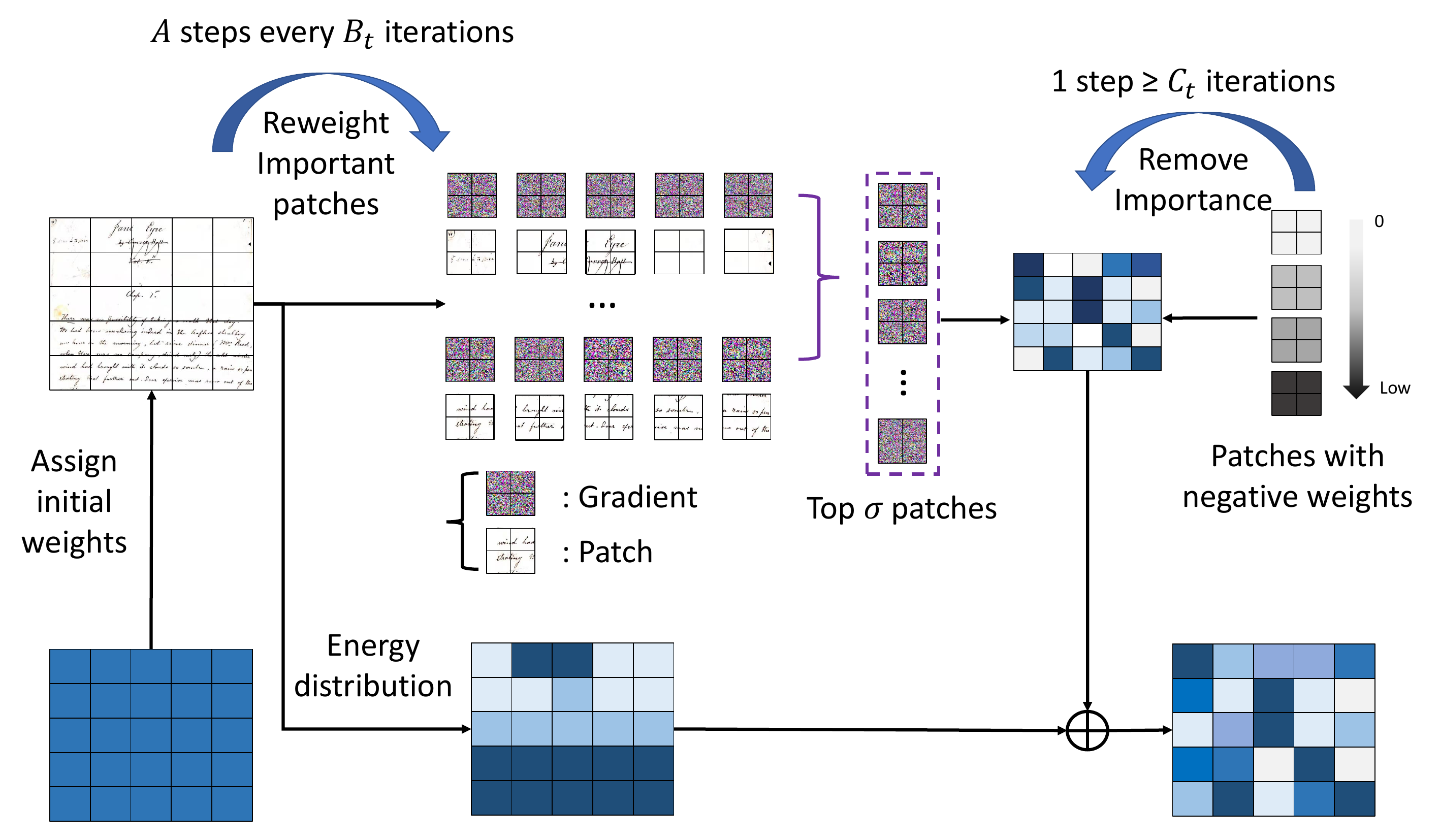}
    \caption{Adaptive matching mechanism. The task-related patches are determined by reweighting all patches based on their contribution towards a correct classification result. In this process, the left steps, i.e., reweight important patches, aim to increase the influence (weight) of important patches, while the right steps, i.e., remove importance, take into account the differences of key patches in homologous augmented samples.}
    \label{fig:Learning patch importance}
\end{figure}

\subsection{Pre-processing}
\label{sec:4.2}
Handwriting classification can be adversely affected by various factors, such as improper storage of paper or external intervention, which can significantly reduce its effectiveness. Common text image defects that affect handwriting quality include scratches, damage, stains, and folds (as shown in Figure \ref{fig:defect}). While manuscripts of famous authors are valuable objects, they still have image defects that can impact the visual quality. For instance, connecting strokes under stain masking can be mistakenly judged as high-density handwriting dots.

To improve the visual quality of manuscripts, we design a plug-and-play energy-oriented operator, denoted as $S_{\phi}$. The operator $S_{\phi}=\sigma \circ R \circ D$ converts handwriting manuscripts into a series of energy distributions, where $\sigma(\cdot)$ denotes the two-dimensional fast Fourier transform, $R(\cdot)$ is the regularization term used to maintain the smoothness, $D(\cdot,\cdot)$ is the data term consisting of similarity constraint. The energy value indicates the probability that it belongs to handwriting features instead of noise. Therefore, filtering out the low-energy pixels can address disturbances caused by handwriting conditions and text preservation in daily life. Figure \ref{fig:defect} (b) shows some samples of authentic works of famous authors collected in our study. 

We start by obtaining the input handwriting data $I=\left \{ \mathcal{I} _{i} \right \}_{i=1}^{N}$, which may contains various noise. Next, we calculate the denoised data $X=\left \{ x_{i} \right \} _{i=1}^N$ with $x_{i}=S_{\phi}(\mathcal{I} _{i})$. We hope that the calculated denoised data $X=S_{\phi}(I)$ can minimize the pre-training loss $\mathcal{L}$ (as mentioned in Eq.\ref{equ:loss}). Then, we fix the model $f_{\theta}$ in the pre-training stage, only update the energy-oriented operator$S_{\phi}$ during pre-processing by:
\begin{equation}
\begin{array}{l}
S_{\phi}\gets S_{\phi}-\beta \nabla _{S_{\phi}}\sum_{\mathcal{I} _{i}\in I}  \mathcal{L} (f_{\theta};S_{\phi}(\mathcal{I} _{i}))  \\[8pt]
\end{array}
\end{equation}
where $\beta$ is the learning rate. The visual quality improvement of manuscripts after pre-processing is shown in Figure \ref{fig:defect} (c). Note that this operator can be introduced into any model to obtain high-quality images.

\subsection{Generalized Pre-training}
\label{sec:4.3}
During this stage, we pre-train the model $f_{\theta}$ through momentum-based adaptive contrastive learning to learn general representations from the pre-processed handwriting data with energy distributions. It consists of two branches, i.e., a contrastive learning branch and a momentum branch.

As illustrated in Figure \ref{fig:SherlockNet}, we first augment each denoised images $X=\left \{ x_{1},...,x_{N} \right \} \in \mathbb{R}^{H\times W\times C}$, and split each image into a sequence of $M=H\times W/\mathcal{P}^2 $ patches, denoted by $P=\left \{ p_i \right \}_{i=1}^M\in \mathbb{R}^{\mathcal{P}^2\times C}$. Here, $H$, $W$, and $C=2$ represent the height, width of the image, and number of branches, respectively. Then, we split them into two branches using the contrastive learning branch and the momentum branch to learn.

In the first branch, we use an adaptive matching scheme (as illustrated in Figure \ref{fig:Learning patch importance}) to identify the task-related patches that are helpful for decision-making in the current batch, which results in reweighted patches denoted as $P_{w}$. Specifically, we first assign the initial weights on each patch $p_i$, i.e., each weight of $M$ patches is set to $1/M$. Subsequently, we adjust the weights based on the calculated energy distribution, i.e., normalize weights after assigning values via the score of energy. Then, we use a weight boost of $\gamma $ to top-$\sigma$ patches for $A$ steps based on the current gradient of the model $f_{\theta}$ in every $B_t$ iteration. The gradient calculation is as described in Eq.\ref{equ:loss}. Note that although the augmented samples mostly preserve the same properties \cite{cubuk2019autoaugment}, the key patches may vary. Moreover, setting a consistent weight matrix for homologous augmented samples in a batch can improve computational efficiency, but it may also impair performance. Therefore, in the next step, we remove patches with changes smaller than $1/3$ of the mean change every $C_t$ times until reaching the upper limit of iterations $T_{MAX}$ or the threshold for eliminating patches. After combining the weights with the original weights and then normalizing, we obtain $P_{w}$ with more accurate weights.

In the second branch, we just assign the same initial weights to each patch of the augmented images, and demote it as $P_{a}$.

Next, we flatten and put all patches, i.e., $P_{w}$ and $P_{a}$ to the same feature extractor $g(\cdot)$, obtaining the representations of $P_{w}$ which are denoted as $Z_w=\left \{ z_{w}^{n}|n=1,...,N \right \} $, and the representations of $P_{a}$ which are denoted as $Z_a=\left \{ z_{a}^{n}|n=1,..., N \right \} $. Each element in $Z_w$ and $Z_a$ is a vector with length $L$ and belongs to $\mathbb{R}^D$. We select four structures for the feature extractor: Conv4 \cite{chen2020simple}, Resnet-50 \cite{he2020fragnet}, Vision Transformers like ViT \cite{dosovitskiy2020image}, and hierarchical Transformers like Swin \cite{liu2021swin}. Among them, Conv4, Resnet-50, and ViT satisfy $L=M$, whereas Swin generally produces a reduced number of representations $L<M$ due to internal merging strategies.

After obtaining the representations $Z_{w}$ and $Z_{a}$, we use the modules of the contrastive self-supervised learning branch and the momentum branch \cite{he2020momentum} to perform further learning. In the first branch, we sequentially process the representations $Z_{w}$ through $P^{pat}$, $P^{pro}$, and $P^{pre}$. The patch head $P_{\theta}^{pat}$ projects the extracted representations $Z_{w}$ into a single vector $p_{w}^{pat}$. The projection head $P^{pro}$ and prediction head $P^{pre}$ consist of a 3-layer MLP and a 2-layer MLP, respectively, to further process $p_{w}^{pat}$. Each layer of the MLP includes a fully-connected layer and a normalization layer. Then, we obtain the embeddings of the training samples in the current batch, denoted as $p_{w}^{pre}$. In the momentum branch, we use a patch head $K^{pat}$ and a projection head $K^{pro}$, which have the same architecture as the contrastive learning branch, to process $Z_{a}$. In this branch, we obtain the embeddings denoted as $k_{a}^{pre}$.

After obtaining the embeddings $p_{w}^{pre}$ and $k_{a}^{pre}$, we establish semantic correspondences and broadcast them to compute the contrastive loss, further updating the model $f_{\theta}$. The learning objective can be expressed as:
\begin{equation}
\label{equ:loss}
\begin{array}{l}
    f_{\theta}\gets f_{\theta}-\alpha  \nabla_{f_{\theta}} \sum_{\mathcal{X}_{cl},\mathcal{X}_{mo}\in X} \mathcal{L} (f_{\theta};X) \\[8pt]
    s.t.\quad \mathcal{L} (f_{\theta};X) = \lambda _1\mathcal{L}_{cl} (f_{\theta};\mathcal{X}_{cl})+\lambda _2\mathcal{L}_{mo} (f_{\theta};\mathcal{X}_{mo})\\[8pt]
\end{array}
\end{equation}
where $\alpha$ denotes the learning rate, $\mathcal{X}_{cl}$ and $\mathcal{X}_{mo}$ denote the training data of the two different channels, $\lambda _1$ and $\lambda _2$ are the importance of two different branch losses, i.e., loss of the contrastive learning branch $\mathcal{L}_{cl}$ and the loss of the momentum branch $\mathcal{L}_{mo}$, in the overall optimization loss $\mathcal{L}$. Both $\mathcal{L}_{cl}$ and $\mathcal{L}_{mo}$ are the infoNCE loss as mentioned in Eq.\ref{eq:infonce} but with different embeddings $p_{w}^{pre}$ and $k_{a}^{pre}$ in different branches.

\subsection{Personalized Fine-tuning}
\label{sec:4.4}
During the personalized fine-tuning stage, SherlockNet aims to adapt well to downstream tasks based on small amounts of labeled data $\mathcal{D}^{ds}_{tr}=(I^{ds}_{tr}, Y^{ds}_{tr}) $, where $I^{ds}_{tr}$ and $Y^{ds}_{tr}$ denote the samples and the corresponding labels, respectively. We fine-tune the model $f_{\theta}$ by freezing the trainable weights in the feature extractor, and training a coefficient matrix $\mathcal{W}$ which is followed by a linear layer to predict the writer's identity. Since the handwriting data has no inherent order relationship, we first randomly select non-repetitive training data for fine-tuning in one batch. Then, we use the trained and fixed energy-oriented operator to obtain denoised data $X_{tr}^{ds}$. Next, we perform fine-tuning, and the objective can be expressed as:
\begin{equation}
\label{eq:fine-tuning}
\begin{array}{l}
\arg \min_{\mathcal{W}} \mathcal{L}_{\mathrm{ds}}(\mathcal{W};f_\theta, \mathcal{D}^{ds})  \\[8pt]
s.t. \quad \mathcal{L}_{\mathrm{ds}}(\mathcal{W};f_\theta, \mathcal{D}^{ds})=-\sum_{{x_i^{ds}}\in x}\log P(y^{ds}_{i}; \mathcal{W} f_{\theta}(x^{ds}_{i}))
\end{array}
\end{equation}
where $x_i^{ds}$ and $y_i^{ds}$ are the training sample and the corresponding label. The fine-tuning process only needs a few optimization steps to achieve great adaptation.

\subsection{Practical Application}
\label{sec:4.5}

During the practical application stage, SherlockNet aims to be applied well in the real world conveniently even with damaged or forged handwriting data. This stage is user-oriented and mainly includes two parts: (i) evaluating the effect of the model on downstream tasks, and (ii) designing convenient application interfaces to facilitate deployment in reality. To verify the effectiveness of the model, we use the test samples of the downstream task, represented as $\mathcal{D}^{ds}_{te}= (I^{ds}_{te}, Y^{ds}_{te})$. The performance of the model $f_{\theta}$ with the fine-tuned coefficient matrix $\mathcal{W}$ on the downstream task can be expressed as:
\begin{equation}
\label{eq:performance}
\begin{array}{l}
\mathcal{R}(\mathcal{W} ;f_\theta, \mathcal{D}^{ds}_{te}) = \mathbb{E}_{(\mathcal{I} ,y) \sim \mathcal{D}^{ds}_{te}}[-\log p(y ; \mathcal{W}  f_{\theta}(S(\mathcal{I})))] \\\nonumber
\end{array}
\end{equation}
where $\mathcal{R}$ denotes the performance risk on the downstream task. 

For ease of use in real-world applications, we adopt a modular architecture in SherlockNet following \cite{li2013integrated, sullivan2001structure}: (i) Model modularization: we decompose SherlockNet into multiple modules, such as energy-oriented operator, contrastive learning branch, and momentum branch. This allows users to activate or deactivate each module as needed. (ii) API layering: We design multiple levels of API, each corresponding to a different stage, completing different tasks. For example, the high-level API is used to complete the entire verification process at once, while the low-level API allows users to call different parts of the model separately.

 %%%%%%%%%%%%%%%%%%%%%%%%%%%%%%%%%%%%%%%%%%%%%%%%%%%%%%%%%%%%%%%%%%%%
 %%%%%%%%%%%%%%%%%%%%%%%%%%%%%%%%%%%%%%%%%%%%%%%%%%%%%%%%%%%%%%%%%%%%
 
%% -----------------Empirical analysis-----------------------

\section{Experiments}
\label{sec:5}
In this section, we first introduce the experimental settings in Subsection \ref{sec:6.1}. Then, we conduct comparative experiments between SherlockNet and the state-of-the-art (SOTA) baselines from four aspects: performance, clustering effect, model size, and inference time in Subsection \ref{sec:6.2}. Next, we simulate realistic scenarios such as data corruption and forgery in Subsection \ref{sec:6.3} to analyze the robustness of our approach. Furthermore, we perform ablation study and visualization analyses in Subsections \ref{sec:6.4} and \ref{sec:6.5} respectively to explore how our SherlockNet works. Finally, we hold further discussions and plan our future works in Subsection \ref{sec:6.6}.

\subsection{Experimental Settings}
\label{sec:6.1}
In this subsection, we introduce the datasets, baselines, and implementation details of our experiments in sequence.

\begin{figure*}
    \centering
    \includegraphics[width=0.9\textwidth]{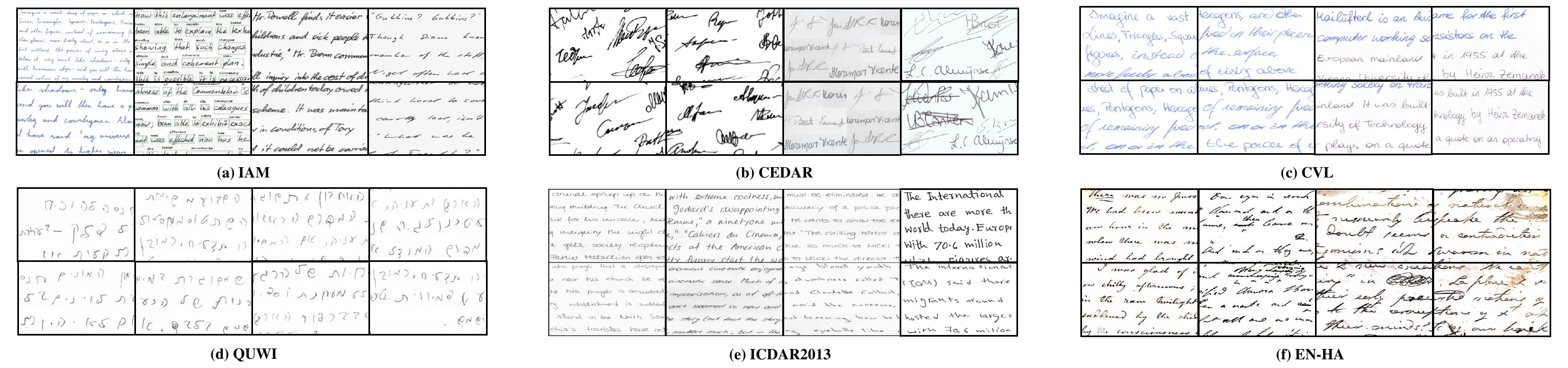}
    \caption{Examples of the six benchmark datasets used in the experiments, including IAM \cite{marti1999full}, CEDAR \cite{srihari2002individuality}, CVL \cite{kleber2013cvl}, QUWI \cite{al2012quwi}, ICDAR2013 \cite{hassaine2013icdar}, and our own constructed dataset EN-HA.}
    \label{fig:5.1}
\end{figure*}

\subsubsection{\textbf{Datasets}}
Our SherlockNet is evaluated on six benchmark datasets, i.e., IAM, CEDAR, CVL, QUWI, ICDAR2013, and a novel dataset called EN-HA. Note that we reconstruct the datasets and use randomly spliced handwriting pages as training data to simulate situations that exist in real life. Figure \ref{fig:5.1} gives some examples of these six datasets.

IAM \cite{marti1999full} contains 13353 labeled text lines of content written in English by 657 writers, with approximately 20 lines per writer. It includes 1539 forms as labels, providing detailed information such as writer identity. In our experiments, we construct 4 handwriting pages for each author, each of which consists of 5 lines of text randomly selected and spliced together. This reconstructed dataset is called IAM-FHA.

CEDAR \cite{srihari2002individuality} contains 105573 words written in English by 200 writers, with approximately 528 words per writer. In our experiments, we randomly splice 500 words of the same author together to form a handwriting page and construct 20 handwriting pages for each author, denoted as CEDAR-FHA.

CVL \cite{kleber2013cvl} contains 1604 handwriting pages written in both English and German by 310 writers, with 27 of them writing 7 pages and 283 writers writing 5 pages. We crop the handwriting pages so that each sample contains approximately 500 characters. This reconstructed dataset is called CVL-FHA.

QUWI \cite{al2012quwi} contains 4068 handwriting pages written in both Arabic and English by 1017 writers, with 4 pages per writer. We only use the first and third pages in our experiments. Similar to CVL, we crop the handwriting pages, and this reconstructed dataset is called QUWI-FHA.

ICDAR2013 \cite{hassaine2013icdar} contains 1900 handwriting pages written in both Arabic and English by 475 writers. In our experiments, we apply all the pages from each author for evaluation, where the first and second pages are written in Arabic while the third and fourth pages are written in English.

EN-HA is a novel label-free handwriting dataset we collected for freeform handwriting authentication. It contains 800 handwriting pages written in English by 40 writers, with 20 being famous historical figures and 20 being volunteers. The manuscripts of famous historical figures are obtained from the Internet, mainly provided by the British Museum and Vatican Museum. The dataset simulates real-world situations, which consists of 90\% of real-life images (average noisy area around 10\%) and 10\% of fake images.

\subsubsection{\textbf{Baselines}}
We categorize the baselines into three groups: (i) self-supervised methods, (ii) traditional handwriting authentication methods, and (iii) advanced handwriting authentication methods (including SOTA methods for each dataset). Next, we will introduce these three types of baselines for freedom handwriting authentication (FHA).

Self-Supervised Baselines. We select classic SSL baselines to evaluate the advantages of SherlockNet over other SSL frameworks in handwriting authentication, including SimCLR \cite{chen2020simple}, BYOL \cite{grill2020bootstrap}, Barlow Twins \cite{zbontar2021barlow}, and MOCO \cite{he2020momentum}.

Traditional Handwriting Authentication Baselines. When data is scarce or computational resources are limited, traditional methods using handcrafted features may be more robust, as well as easier to explain \cite{ferrari2019hand}. Therefore, we introduce this type of method to evaluate the robustness and scalability of SherlockNet, including NN-LBP/LPQ/LTP \cite{chahi2019effective}, CoHinge/QuadHinge \cite{he2017beyond}, COLD \cite{he2017beyond}, and Chain Code Pairs/Triplets (CC-Pairs/Triplets) \cite{siddiqi2010text}. They use different types of features, including local texture features, joint features based on raw Hinge kernels, line distribution features, and unconstrained handwriting visual features, respectively.

Advanced Handwriting Authentication Baselines. To evaluate the performance of SherlockNet in FHA, we select various advanced and SOTA methods for comparison, including FragNet \cite{he2020fragnet}, GR-RNN \cite{he2021gr}, SEG-WI \cite{kumar2020segmentation}, Siamese-OWI \cite{kumar2022siamese}, Deep-HWI \cite{javidi2020deep}, SWIS \cite{manna2022swis}, SURDS \cite{chattopadhyay2022surds}, SVV-TF \cite{SVV-TF}, CAE-SVM \cite{CAE-SVM}, DeepNet-WI \cite{deepnetWI}, and WriterINet \cite{WriterINet}.

\subsubsection{\textbf{Implementation Details}}
For the feature extractor, we select four types of backbones: Conv4 \cite{chen2020simple}, Resnet-50 \cite{he2020fragnet}, vision Transformers like ViT \cite{dosovitskiy2020image}, and hierarchical Transformers like Swin \cite{liu2021swin}. The Conv4 is adopted as the default backbone with an embedding size of 512. We use a unified backbone structure (Conv4 or Resnet50) in comparison experiments while trying all the above four backbones to find the optimal structure in the ablation study. For adaptive matching, the step size of reweighting the patches is set to 3 ($A=3$) with $\gamma=0.02$, $\sigma$ is set to 10, $B_{t}$ and $C_{t}$ are set to 10 and 20, respectively. The hyperparameters $\lambda_1$ and $\lambda_2$ in the learning objective are set to 0.6 and 0.3, respectively. For optimizer, we use Adam \cite{kingma2014adam} to train models, where Momentum and weight decay are set at $0.9$ and $10^{-4}$, respectively. Other hyperparameters are: The initial learning rates for all experiments are established as $\alpha=0.3$ and $\beta=0.4$, the batch size as 1024, and the weight decay of 0.05. Considering the nature and texture features of handwriting data, we apply various data augmentations including random Gaussian blurring, random mixup, random horizontal flip across clips, etc. All experiments are conducted with NVIDIA V100 GPUs and NVIDIA 4090ti GPUs.

%% ---Table 1---
\begin{table*}
  \centering
  \caption{Accuracy(\%) of different methods on six benchmark datasets. The optimal results are highlighted in \textbf{bold}. Note that we use randomly spliced handwriting pages as training data and use Resnet-50 as the backbone which is different from the original settings. Therefore, the performance may differ from that in papers, e.g., FragNet from $79.8\%$ to $69.8\%$.}
  \label{tab:comparison}
  \resizebox{0.9\linewidth}{!}{
  \begin{tabular}{l|cc|cc|cc|cc|cc|cc}
    \toprule
    \multirow{2}{*}{\textbf{Methods}} & \multicolumn{2}{c|}{\textbf{IAM-FHA}} & \multicolumn{2}{c|}{\textbf{CEDAR-FHA}} & \multicolumn{2}{c|}{\textbf{CVL-FHA}} & \multicolumn{2}{c|}{\textbf{QUWI-FHA}} & \multicolumn{2}{c|}{\textbf{ICDAR2013-FHA}} & \multicolumn{2}{c}{\textbf{EN-HA}}\\
    % \cline{2-13}
    & \textbf{Top 1} & \textbf{Top 5} & \textbf{Top 1} & \textbf{Top 5} & \textbf{Top 1} & \textbf{Top 5} & \textbf{Top 1} & \textbf{Top 5} & \textbf{Top 1} & \textbf{Top 5} & \textbf{Top 1} & \textbf{Top 5} \\
    \midrule
    \textbf{SimCLR} \cite{chen2020simple} & 61.412 & 80.515 & 75.832 & 82.012 & 62.301 & 85.298 & 49.785 & 56.353 & 51.936 & 69.501 & 45.893 & 66.289 \\
    \textbf{BYOL} \cite{grill2020bootstrap} & 53.238 & 79.789 & 71.520 & 79.233 & 49.872 & 82.011 & 50.542 & 68.487 & 55.293 & 71.220 & 44.732 & 71.458 \\
    \textbf{Barlow Twins} \cite{zbontar2021barlow} & 49.947 & 86.289 & 71.389 & 79.278 & 62.332 & 80.299 & 42.891 & 57.777 & 50.829 & 68.839 & 48.203 & 68.128 \\
    \textbf{MOCO} \cite{he2020momentum} & 64.852 & 82.545 & 69.298 & 80.122 & 58.825 & 71.513 & 53.044 & 61.825 & 49.063 & 74.202 & 56.256 & 77.197 \\
    \midrule
    \textbf{NN-LBP} \cite{chahi2019effective} & 18.512 & 31.293 & 24.355 & 39.053 & 13.523 & 28.238 & 9.328 & 17.938 & 19.544 & 35.205 & 10.083 & 21.279 \\
    \textbf{NN-LPQ} \cite{chahi2019effective} & 18.148 & 32.932 & 25.534 & 37.562 & 14.200 & 30.856 & 10.254 & 17.652 & 17.025 & 34.545 & 12.877 & 22.830 \\
    \textbf{NN-LTP} \cite{chahi2019effective} & 17.843 & 29.842 & 24.378 & 37.234 & 14.784 & 30.239 & 9.010 & 16.382 & 21.019 & 38.231 & 11.793 & 24.873 \\
    \textbf{CoHinge} \cite{he2017beyond} & 19.622  & 35.215 & 40.420 & 51.527 & 18.164 & 34.055 & 15.058 & 22.024 & 22.027 & 44.789 & 14.109 & 26.724 \\
    \textbf{QuadHinge} \cite{he2017beyond} & 20.984 & 36.492 & 40.281 & 52.098 & 16.373 & 37.017 & 15.441 & 25.093 & 25.234 & 44.897 & 15.389 & 27.018 \\
    \textbf{COLD} \cite{he2017beyond} & 11.893 & 27.809 & 39.839 & 48.879 &  17.132 & 35.500 & 13.202 & 20.865 & 20.052 & 39.581 & 10.035 & 25.284 \\
    \textbf{CC-Pairs} \cite{siddiqi2010text} & 13.480 & 27.652 & 30.932 & 48.039 & 19.892 & 30.180 & 12.209 & 24.278 & 27.492 & 41.033 & 20.840 & 45.923 \\
    \textbf{CC-Triplets} \cite{siddiqi2010text} & 15.415 & 34.732 & 34.893 & 51.289 & 19.010 & 31.122 & 12.757 & 25.207 & 28.565 & 44.028 & 19.982 & 49.284 \\
    \midrule
    \textbf{FragNet} \cite{he2020fragnet} & 69.891 & 85.055 & 86.890 & 93.238 & 77.303 & 92.832 & 48.202 & 71.252 & 54.250 & 81.651 & 64.382 & 90.837 \\
    \textbf{GR-RNN} \cite{he2021gr} & 70.235  & 85.724 & 77.242 & 89.559 & 79.541 & 94.466 & 50.605 & 69.366 & 68.798 & 89.387 & 56.852 & 91.015 \\
    \textbf{SEG-WI} \cite{kumar2020segmentation} & 77.308 & 89.952 & 75.190 & 85.192 & 65.227 & 86.692 & 44.065 & 53.251 & 62.588 & 84.524 & 71.897 & 84.240 \\
    \textbf{Siamese-OWI} \cite{kumar2022siamese} & 81.978 & 92.387 & 84.527 & 92.789 & 50.132 & 89.892 & 55.798 & 70.787 & 70.673 & 86.398 & 72.190 & 93.265 \\
    \textbf{Deep-HWI} \cite{javidi2020deep} & 79.720 & 93.516 & 79.254 & 90.865 & 58.527 & 90.011 & 60.522 & 71.232 & 65.386 & 90.524 & 54.350 & 75.085 \\ 
    \textbf{SWIS} \cite{manna2022swis} & 64.587 & 80.890 & 86.892 & 94.110 & 49.897 & 82.524 & 40.522 & 62.252 & 56.832 & 73.812 & 55.132 & 76.798 \\ 
    \textbf{SURDS} \cite{chattopadhyay2022surds} & 60.541 & 84.205 & 73.192 & 89.182 & 67.027 & 86.821 & 44.425 & 58.633 & 50.562 & 78.659 & 67.852 & 92.636\\ 
    \textbf{SVV-TF} \cite{SVV-TF} & 75.156 & 92.001 & 85.911 & 92.150 & 74.237 & 91.836 & \textbf{61.028} & 75.651 & 65.851 & 89.202 & 72.063 & 91.522 \\
    \textbf{CAE-SVM} \cite{CAE-SVM} & 71.250 & 86.530 & 83.601 & 89.845 & 71.122 & 85.137 & 56.890 & 68.135 & 62.879 & 72.510 & 70.569 & 89.622 \\
    \textbf{DeepNetWI} \cite{deepnetWI} & 74.055 & 90.287 & 81.109 & 92.892 & 70.120 & 88.808 & 54.244 & 68.267 & 65.725 & 74.527 & 71.180 & 89.022 \\
    \textbf{WriterINet} \cite{WriterINet} & 76.784 & 92.983 & 82.100 & 93.288 & 69.231 & 83.656 & 60.112 & 72.470 & 66.652 & 74.451 & 71.419 & 90.389 \\
    \midrule
    \textbf{SherlockNet (Ours)} & \textbf{83.290} & \textbf{96.101} & \textbf{87.653} & \textbf{95.019} & \textbf{81.004} & \textbf{95.129} & 60.437 & \textbf{76.748} & \textbf{71.178} & \textbf{91.908} & \textbf{82.782} & \textbf{94.355} \\
    \bottomrule
  \end{tabular}}
\end{table*}

%% ---Table 2----
\begin{table*}
  \centering
  \caption{Accuracy(\%) on EN-HA with different ratios of defects and fake samples. The ``\textbf{+}" indicates the proportion of supplemented defect areas added to the original data or the proportion of falsified data.}
  \label{tab:robustness}
  \begin{tabular}{l|c|ccc|ccc}
    \toprule
    \multirow{2}{*}{\textbf{Methods}} & \multirow{2}{*}{\textbf{Original}} & \multicolumn{3}{c|}{\textbf{Damaged Data}} & \multicolumn{3}{c}{\textbf{Falsified Data}} \\
    % \cline{3-8}
    & &  \textbf{+10\%} & \textbf{+30\%} & \textbf{+50\%} & \textbf{+10\%} & \textbf{+20\%} & \textbf{+30\%}\\
    \midrule
    \textbf{SimCLR} \cite{chen2020simple} & 45.83 $\pm $ 0.15 & 44.58 $\pm $ 0.28 & 42.02 $\pm $ 0.21 & 37.03 $\pm $ 0.22 & 42.20 $\pm $ 0.25 & 36.52 $\pm $ 0.13 & 29.65 $\pm $ 0.29 \\
    \textbf{BYOL} \cite{grill2020bootstrap} & 44.72 $\pm $ 0.21 & 44.03 $\pm $ 0.15 & 42.12 $\pm $ 0.18 & 40.35$\pm $0.09 & 42.42 $\pm $ 0.26 & 38.05 $\pm $ 0.14 & 30.51 $\pm $ 0.13 \\
    \textbf{MOCO} \cite{he2020momentum} & 56.26 $\pm $ 0.26 & 55.89 $\pm $ 0.20 & 54.81 $\pm $ 0.09 & 52.12 $\pm $ 0.13 & 54.50 $\pm $ 0.18 & 51.50 $\pm $ 0.26 & 46.52 $\pm $ 0.16 \\
    \midrule
    \textbf{FragNet} \cite{he2020fragnet} & 64.38 $\pm $ 0.34 & 64.01 $\pm $ 0.16 & 63.36 $\pm $ 0.25 & 63.57 $\pm $ 0.17 & 63.52 $\pm $ 0.10 & 61.32 $\pm $ 0.22 & 54.61 $\pm $ 0.23 \\
    \textbf{GR-RNN} \cite{he2021gr} & 56.82 $\pm $ 0.11 & 55.79 $\pm $ 0.12 & 53.20 $\pm $ 0.20 & 49.58 $\pm $ 0.32 & 53.72 $\pm $ 0.45 & 46.20 $\pm $ 0.25 & 38.56 $\pm $ 0.31 \\
    \textbf{SEG-WI} \cite{kumar2020segmentation} & 71.87 $\pm $ 0.17 & 71.17 $\pm $ 0.18 & 70.01 $\pm $ 0.09 & 70.17 $\pm $ 0.16 & 69.01 $\pm $ 0.93 & 66.78 $\pm $ 0.37 & 59.53 $\pm $ 0.34 \\
    \textbf{Siamese-OWI} \cite{kumar2022siamese} & 72.19 $\pm $ 0.24 & 72.56 $\pm $ 0.57 & 70.52 $\pm $ 0.11 & 67.36 $\pm $ 0.16 & 69.47 $\pm $ 0.18 & 60.02 $\pm $ 0.08 & 55.89 $\pm $ 0.27 \\
    \textbf{Deep-HWI} \cite{javidi2020deep} & 54.35 $\pm $ 0.46 & 53.63 $\pm $ 0.14 & 54.19 $\pm $ 0.38 & 53.12 $\pm $ 0.09 & 50.09 $\pm $ 0.13 & 46.45 $\pm $ 0.26 & 41.65 $\pm $ 0.15 \\ 
    \textbf{SWIS} \cite{manna2022swis} & 55.13 $\pm $ 0.21 & 54.28 $\pm $ 0.14 & 52.12 $\pm $ 0.09 & 49.71 $\pm $ 0.13 & 50.89 $\pm $ 0.10 & 43.68 $\pm $ 0.23 & 31.68 $\pm $ 0.36 \\ 
    \textbf{SURDS} \cite{chattopadhyay2022surds} & 67.82 $\pm $ 0.39 & 67.19 $\pm $ 0.71 & 66.82 $\pm $ 0.14 & 64.12 $\pm $ 0.24 & 66.12 $\pm $ 0.18 & 67.13 $\pm $ 0.09 & 65.17 $\pm $ 0.28 \\
    \midrule
    \textbf{SherlockNet (Ours)} & \textbf{82.78 $\pm $ 0.33} & \textbf{82.89 $\pm $ 0.24} & \textbf{81.93 $\pm $ 0.15} & \textbf{81.07 $\pm $ 0.38} & \textbf{82.29 $\pm $ 0.49} & \textbf{81.37 $\pm $ 0.42} & \textbf{79.45 $\pm $ 0.16} \\
    \bottomrule
  \end{tabular}
\end{table*}

\begin{figure*}
    \centering
    \begin{subfigure}{0.19\textwidth}
        \includegraphics[width=\textwidth]{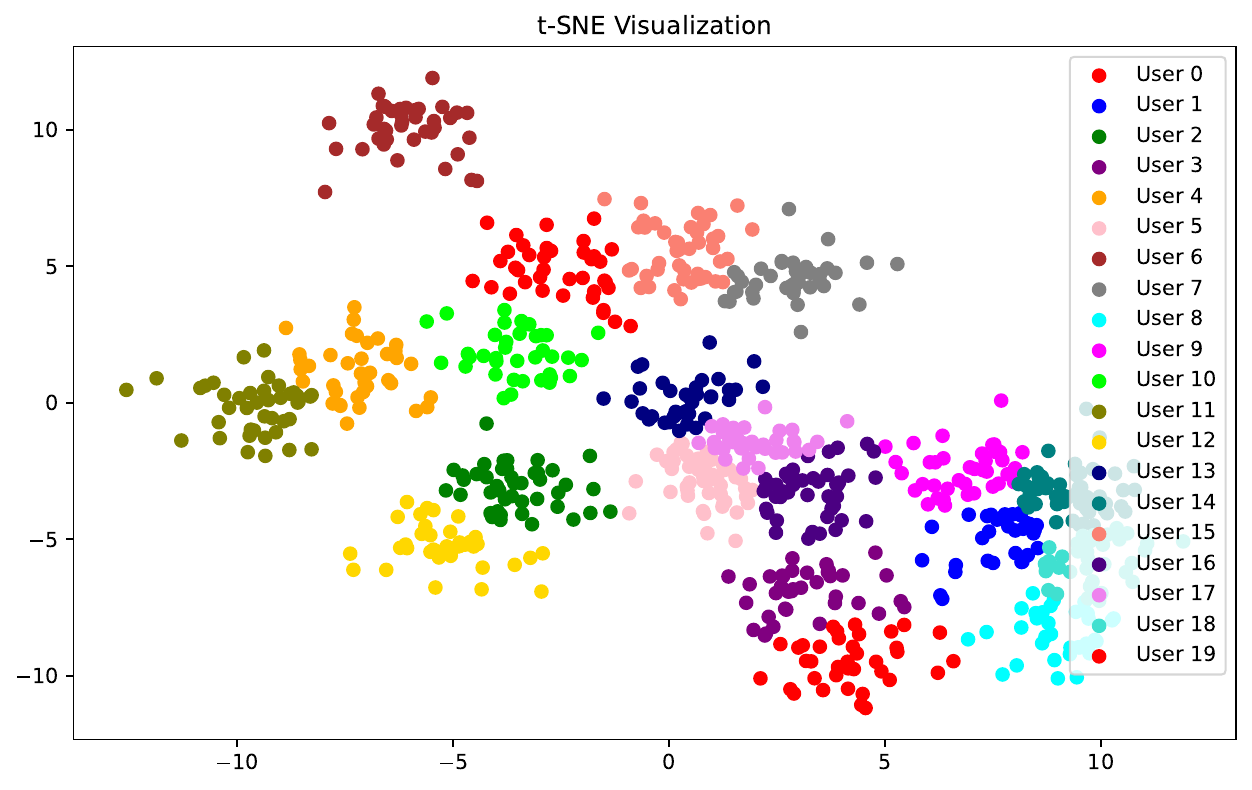}
        \caption{SimCLR}
        \label{fig:sub1}
    \end{subfigure}
    \hfill
    \begin{subfigure}{0.19\textwidth}
        \includegraphics[width=\textwidth]{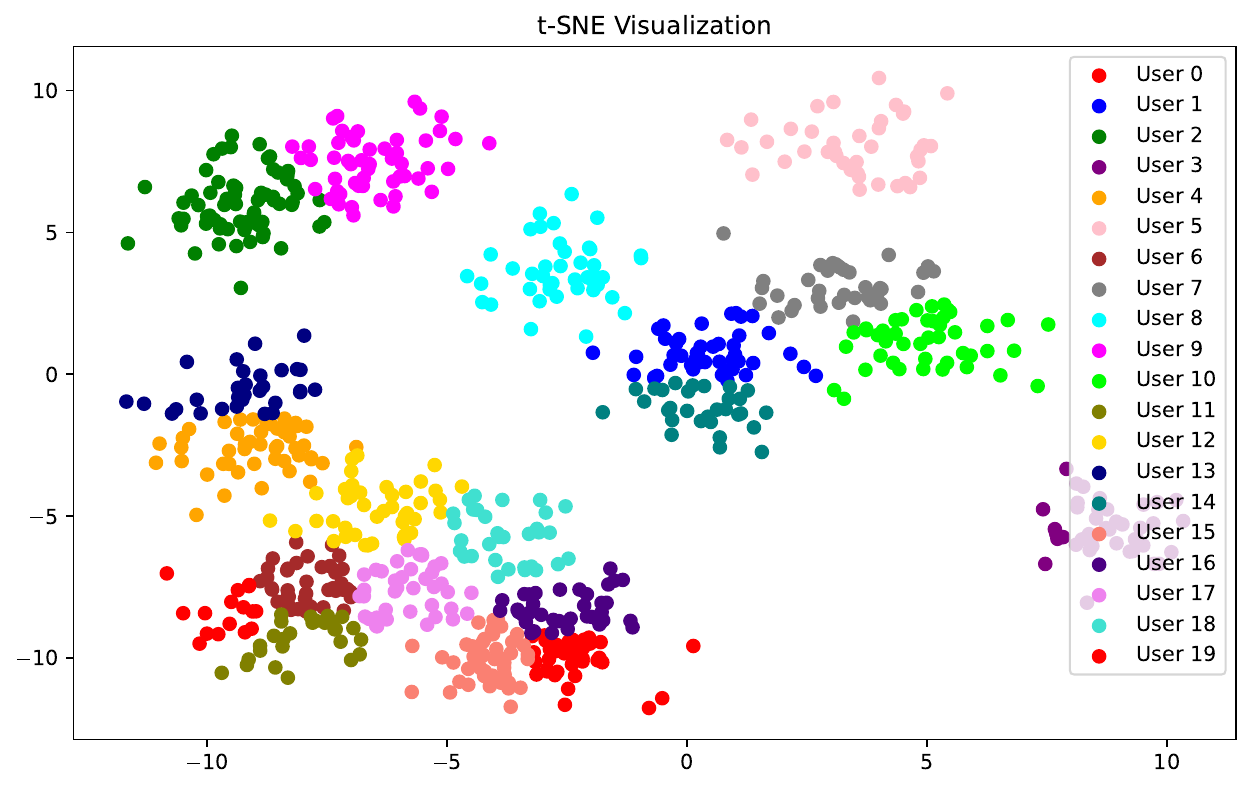}
        \caption{FragNet}
        \label{fig:sub2}
    \end{subfigure}
    \hfill
    \begin{subfigure}{0.19\textwidth}
        \includegraphics[width=\textwidth]{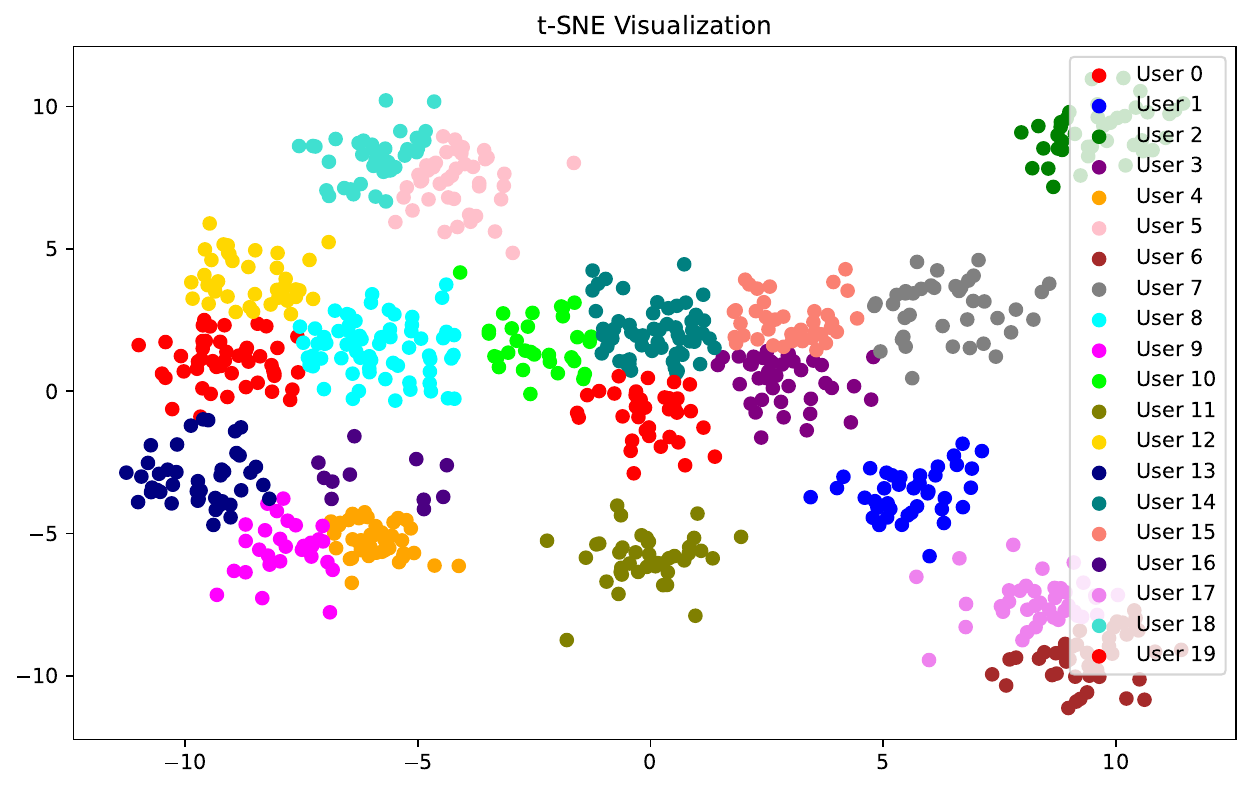}
        \caption{Siamese-OWI}
        \label{fig:sub3}
    \end{subfigure}
    \hfill
    \begin{subfigure}{0.19\textwidth}
        \includegraphics[width=\textwidth]{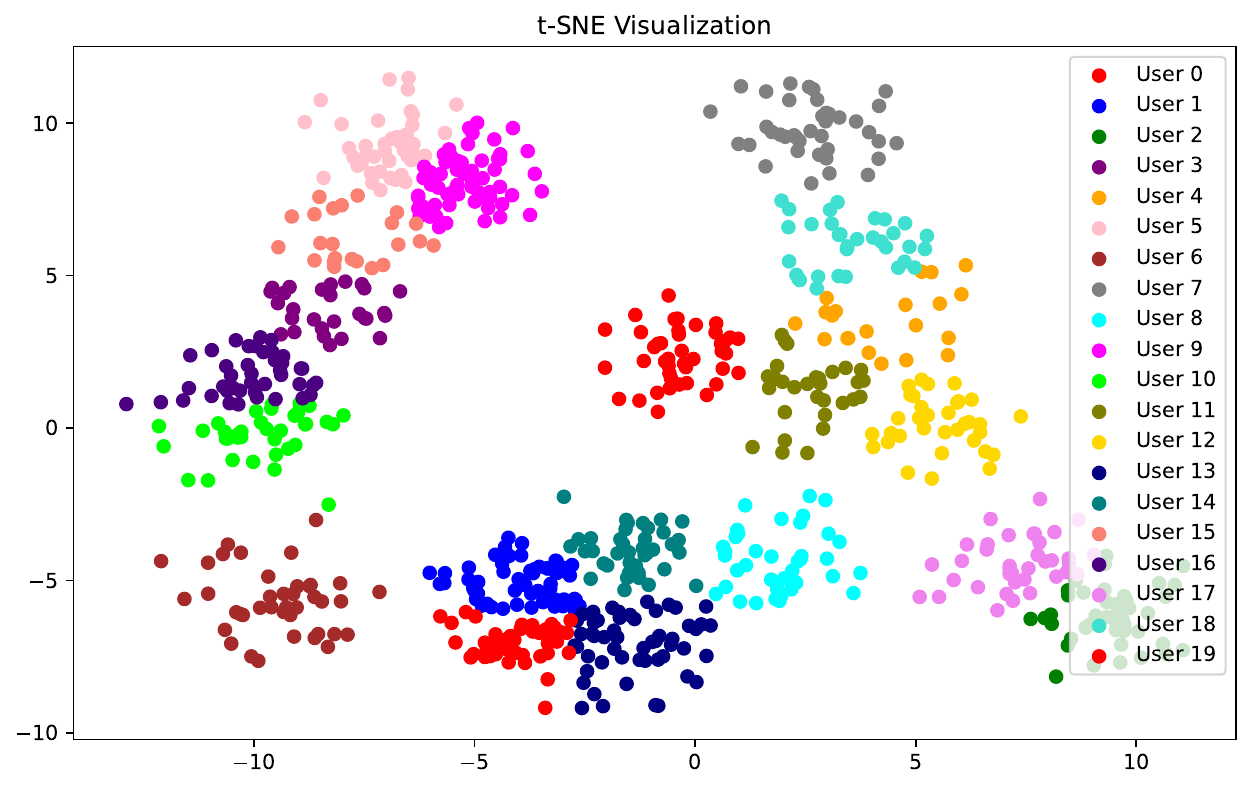}
        \caption{SWIS}
        \label{fig:sub4}
    \end{subfigure}
    \hfill
    \begin{subfigure}{0.19\textwidth}
        \includegraphics[width=\textwidth]{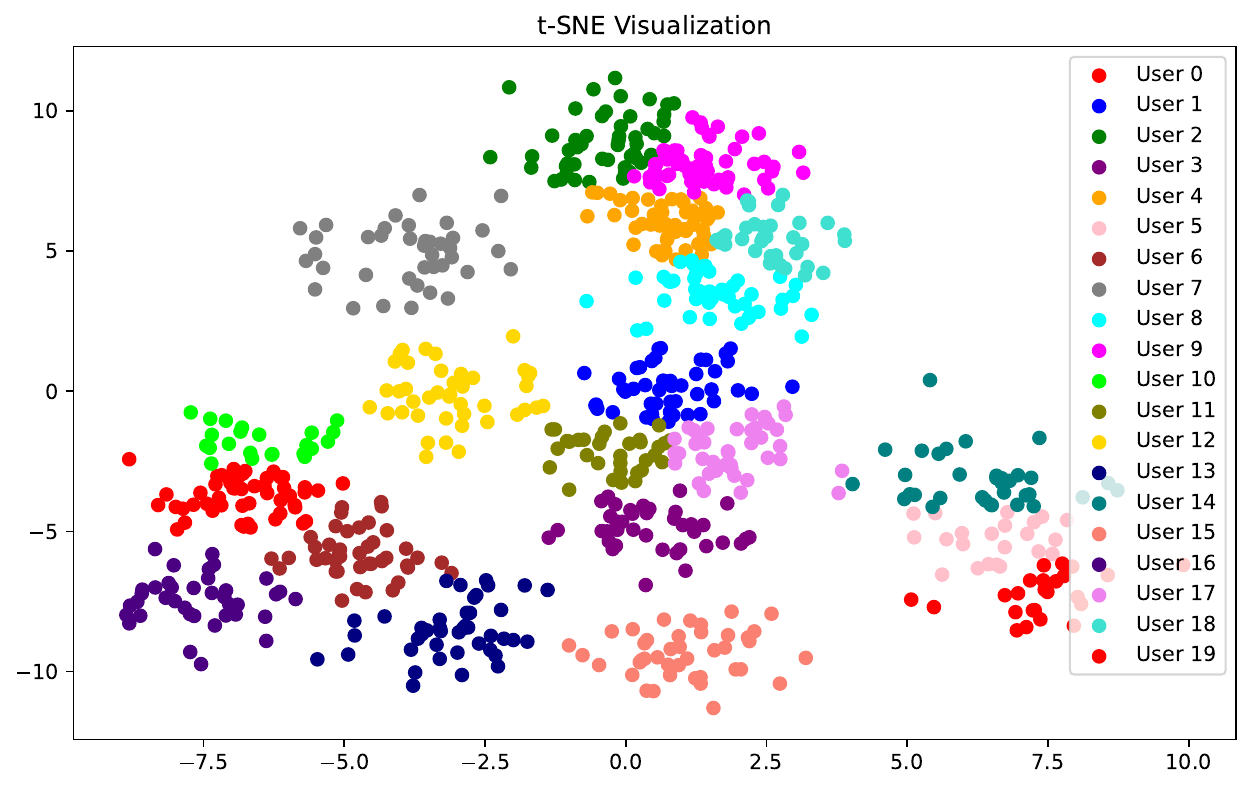}
         \caption{SherlockNet}
        \label{fig:sub5}
    \end{subfigure}
    \caption{t-SNE visualization on IAM-FHA. We randomly select samples from twenty writers for testing.}
    \label{fig:5.2.1}
\end{figure*}

\begin{figure*}
    \centering
    \begin{subfigure}{0.19\textwidth}
        \includegraphics[width=\textwidth]{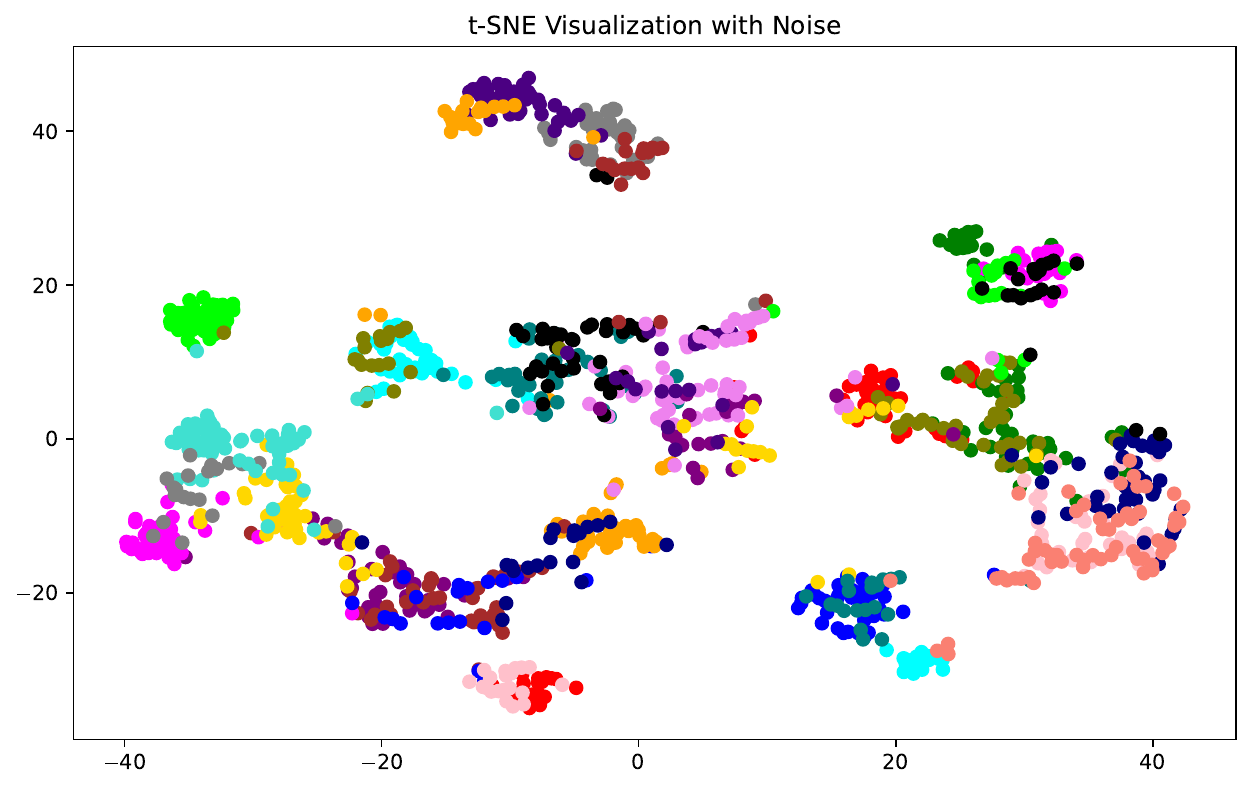}
        \caption{SimCLR}
    \end{subfigure}
    \hfill
    \begin{subfigure}{0.19\textwidth}
        \includegraphics[width=\textwidth]{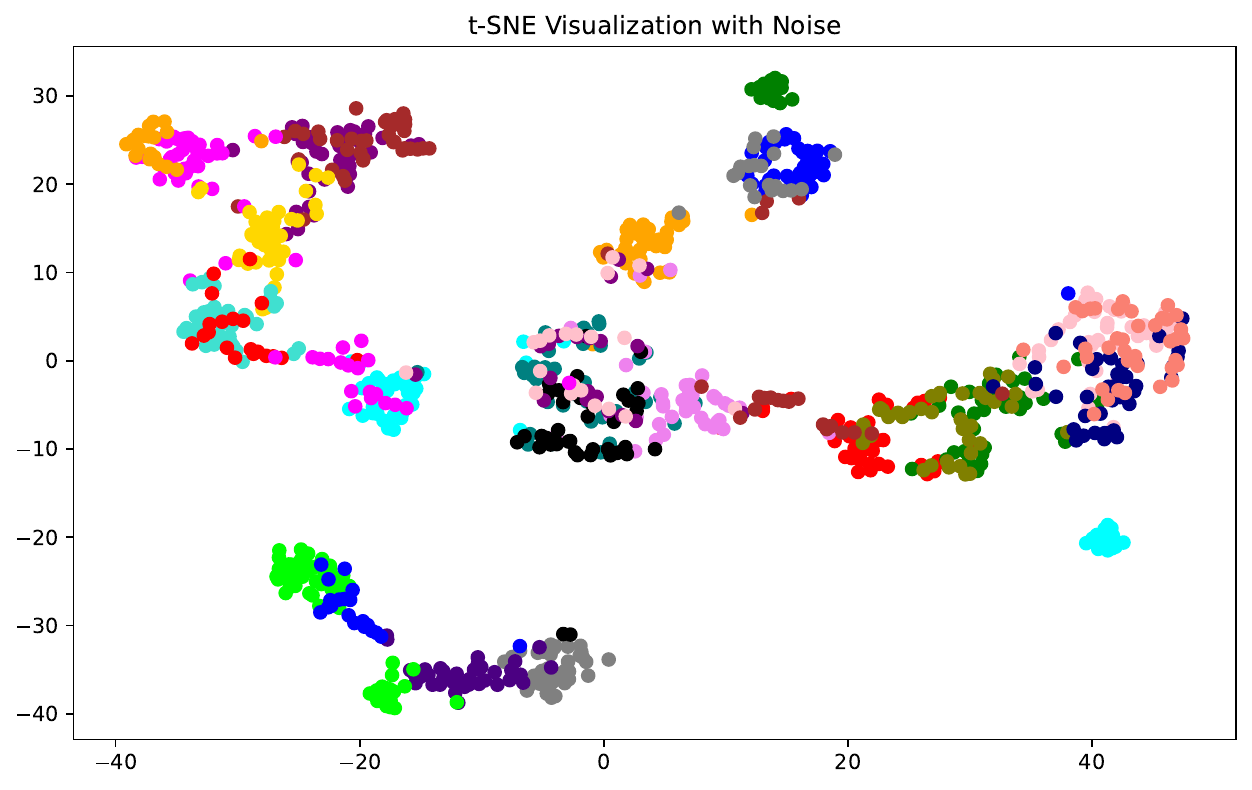}
        \caption{FragNet}
    \end{subfigure}
    \hfill
    \begin{subfigure}{0.19\textwidth}
        \includegraphics[width=\textwidth]{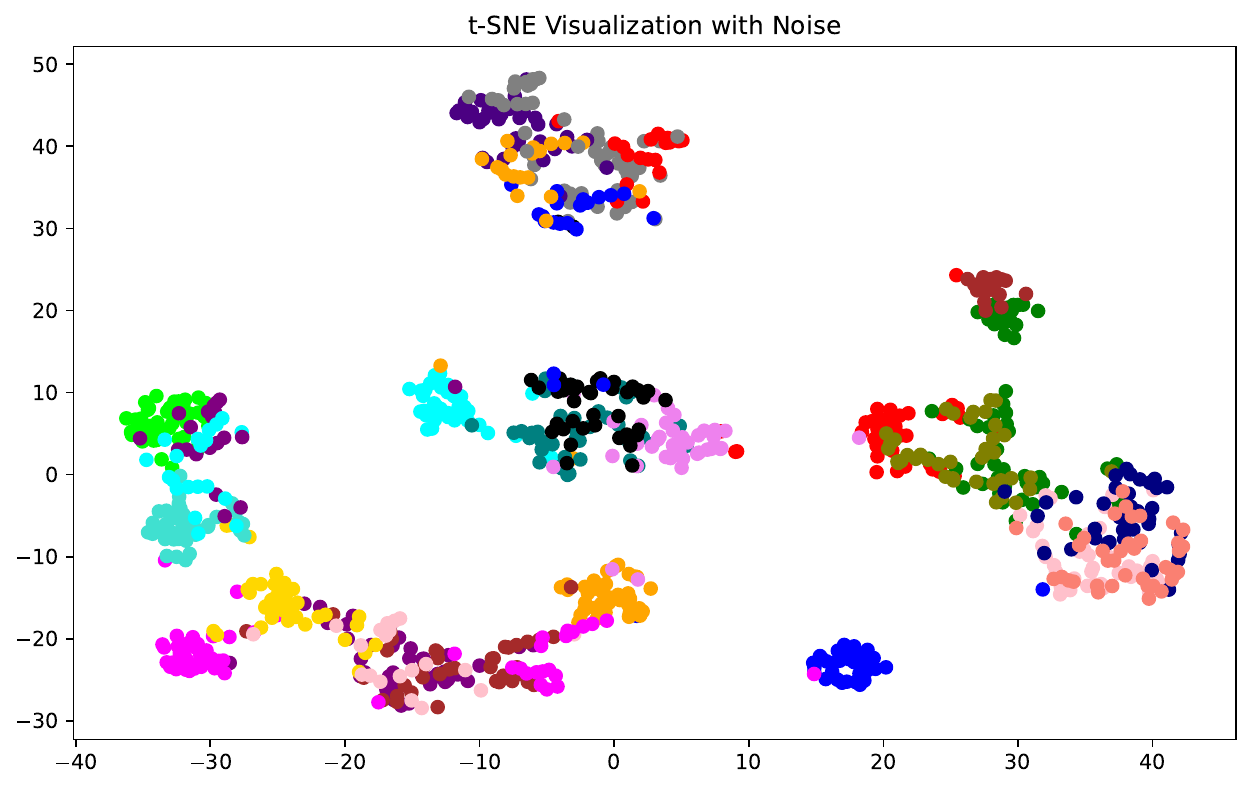}
        \caption{Siamese-OWI}
    \end{subfigure}
    \hfill
    \begin{subfigure}{0.19\textwidth}
        \includegraphics[width=\textwidth]{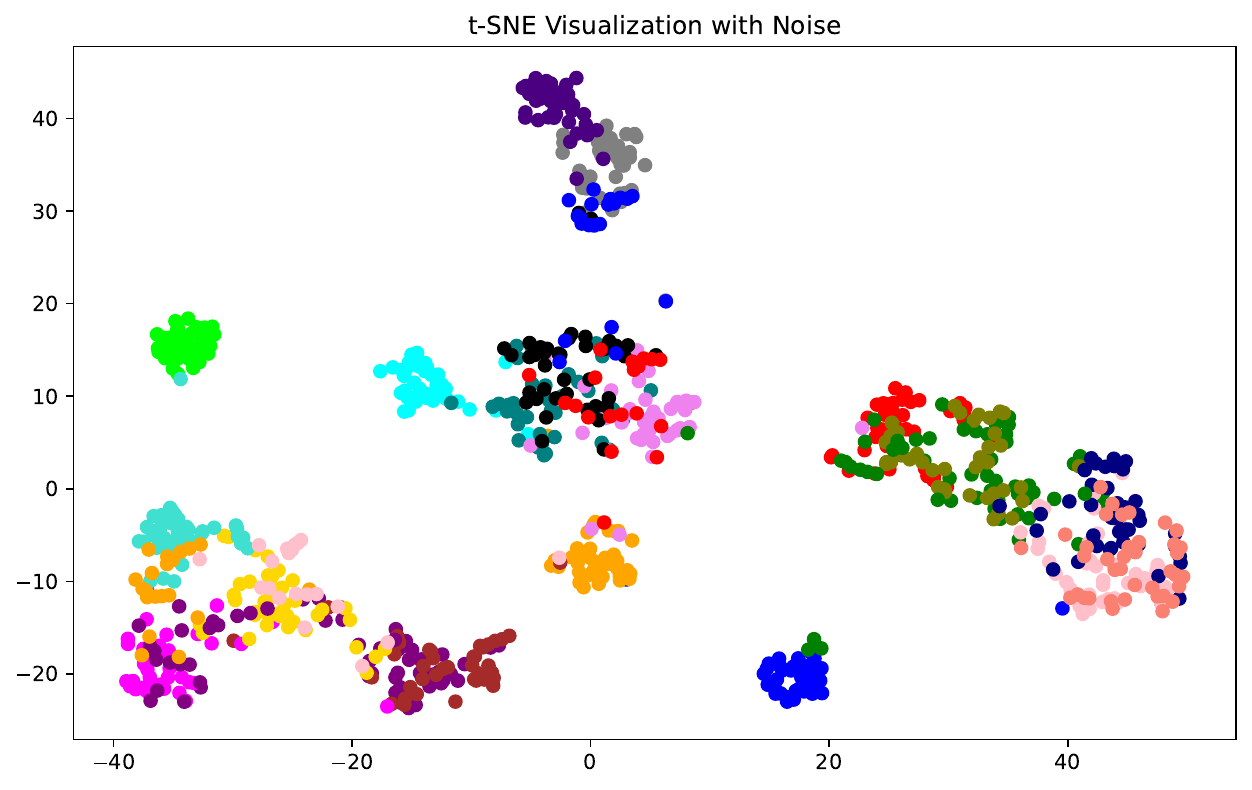}
        \caption{SWIS}
    \end{subfigure}
    \hfill
    \begin{subfigure}{0.19\textwidth}
        \includegraphics[width=\textwidth]{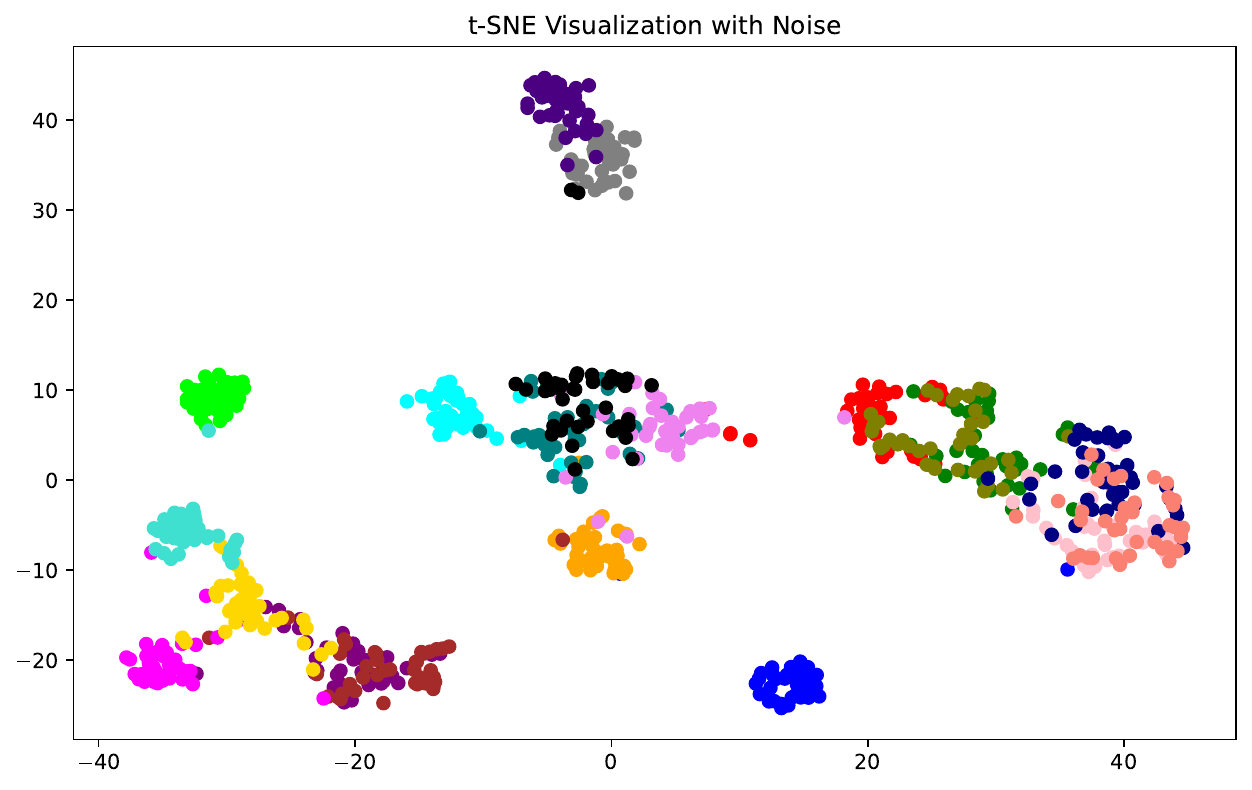}
        \caption{SherlockNet}
    \end{subfigure}
    \caption{t-SNE visualization with noise on EN-HA. We randomly select data from twenty writers for testing, and the data contains various noises. The data points of noise are marked in black.}
    \label{fig:5.2.2}
\end{figure*}

\subsection{Comparison With Baselines}
\label{sec:6.2}
In this subsection, we introduce the comparative experiments conducted to evaluate the effectiveness of the proposed SherlockNet, including comparisons from four perspectives: performance, clustering effect, model size, and inference time. All experiments are performed based on the experimental setup mentioned in Subsection \ref{sec:6.1}.

\subsubsection{\textbf{Quantitative Comparisons}}
\label{sec:6.2.1}
We conduct quantitative comparisons on six benchmark datasets as described in Subsection \ref{sec:6.1}. We reconstruct the datasets, using the handwriting pages as input, and record the average Top 1 and Top 5 classification accuracies of SherlockNet and all baselines.

The results are illustrated in Table \ref{tab:comparison}. Our SherlockNet achieves excellent performance on all benchmark datasets, surpassing almost all SOTA methods. This breakthrough is accomplished without annotated data and observed in almost every dataset, especially in EN-HA, which simulates real-life data forgery and corruption. This achievement underscores the effectiveness and robustness of SherlockNet.

Through further analysis, we obtain three observations: (i) From features: deep learning features perform better than manual features, e.g., the results of FragNet are better than NN-LBP. Learning based on automatic features, e.g., GR-RNN, is more effective than local texture features, e.g., CoHinge, while structural features, e.g., COLD, reduce the recognition effect; (ii) From datasets: image forgery hurts most models, resulting in the Top 1 result of baseline models on EN-HA being far lower than that of SherlockNet, which specifically deals with fake samples. The narrowing of the gap in the Top 5 results indicates that balancing samples and finding key areas are key factors for improving performance; and (iii) From the learning paradigm: our SherlockNet achieves significant progress in SSL frameworks, indicating that the complexity and specificity of handwriting features require specific settings to learn.

\subsubsection{\textbf{Visual Comparisons}}
\label{sec:6.2.2}
Self-supervised learning aims to learn representations that are effective for decision-making. To evaluate the model performance, we use t-SNE \cite{wattenberg2016use} to visualize the performance of different models on the same data distribution. It aims to: (i) project the data points onto the latent space, and judge the quality of the feature representations learned by observing their relative distances; (ii) map the outputs of different models to the latent space, comparing the performance of different methods.

As shown in Figure \ref{fig:5.2.1} and Figure \ref{fig:5.2.2}, we visualize the results of different methods. We obtain three observations: (i) Classification effect: The data points of different classes are clearly separated, and the data points of the same class are close to each other with clear classification centers, indicating that the model effectively preserves the category information. (ii) Local structure preservation: The relative distance between the data points is consistent with the original data, indicating that the model preserves the local structure well. (iii) Outlier separation: The model can distinguish the abnormal information in the data, such as damage, stains, etc. By comparing the t-SNE results of different methods, we find that SherlockNet is better than SOTA baselines in all three aspects.

\subsubsection{\textbf{Model Size and Inference Time Comparisons}}
\label{sec:6.2.3}
Choosing lightweight models and ensuring fast inference are crucial for the performance of specific applications, such as autonomous driving \cite{chen2020denselightnet}, medical diagnosis \cite{ma2020lightweight}, etc. The models that meet the above conditions are more practical in resource-limited environments and facilitate technology deployment. Therefore, we investigate the trade-off between model size and real-time performance. Specifically, we measure the accuracy, model size (MB), and response time (FPS) of different models on a single NVIDIA 4090Ti GPU card. We choose ResNet-50 as the backbone and evaluate the model efficiency on EN-HA.

\begin{figure*}
  \begin{minipage}{0.32\textwidth}
    \centering
    \includegraphics[width=\textwidth]{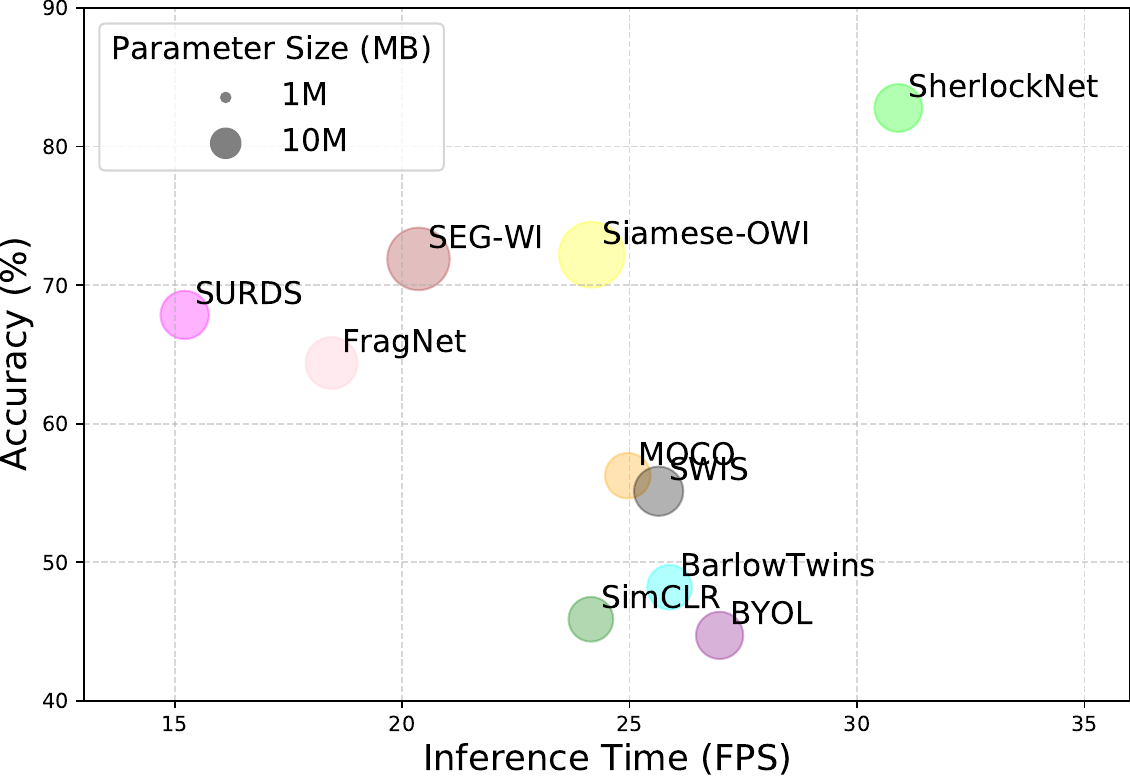}
    \caption{Model efficiency of different models in terms of accuracy, model size, and inference time on EN-HA.}
    \label{fig:5.3}
  \end{minipage}
  \hfill
  \begin{minipage}{0.32\textwidth}
    \centering
    \includegraphics[width=\textwidth]{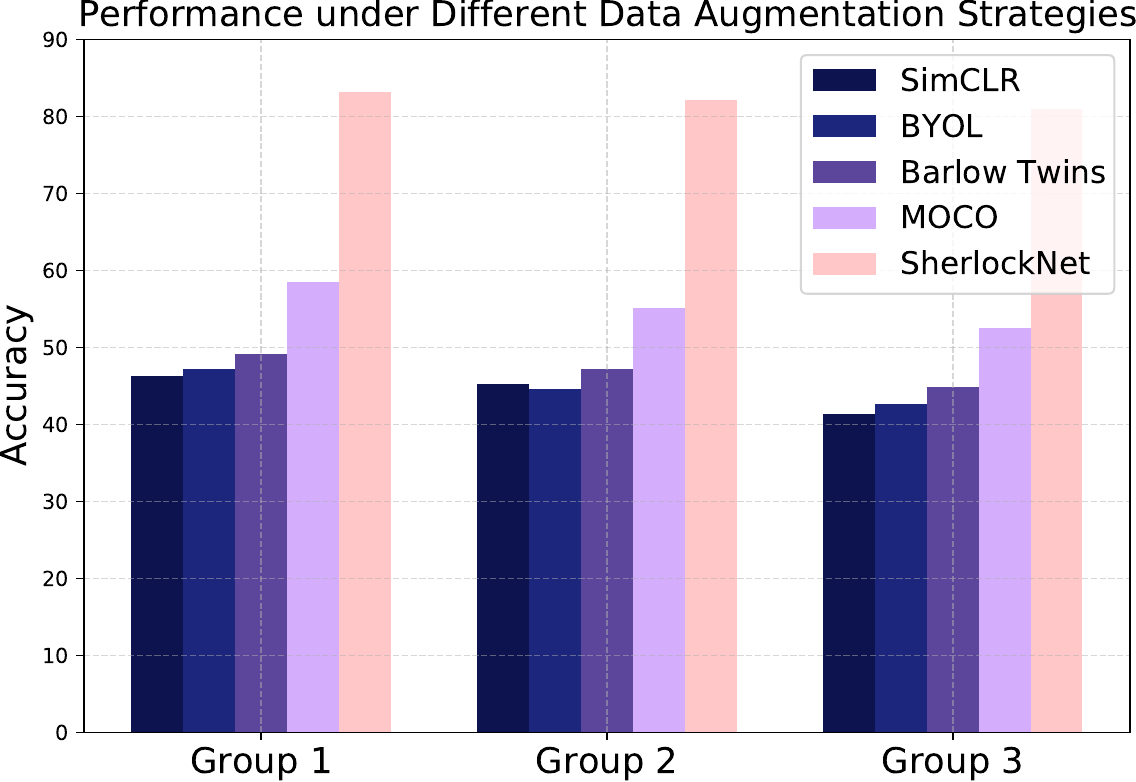}
    \caption{The performance of different models under different data augmentation strategies $A_1\sim A_5$.}
    \label{fig:5.4}
  \end{minipage}
  \hfill
  \begin{minipage}{0.32\textwidth}
    \centering
    \includegraphics[width=\textwidth]{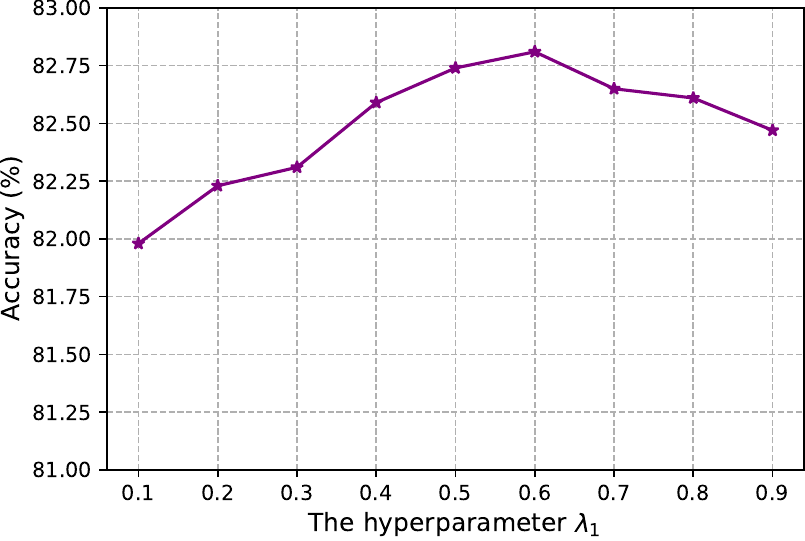}
    \caption{Ablation study of hyperparameters $\lambda_1$ and $\lambda_2$ in SherlockNet, where the horizontal axis is $\lambda_1$, and $\lambda_2=1-\lambda_1$.}
    \label{fig:ablation_2}
  \end{minipage}
\end{figure*}

Figure \ref{fig:5.3} shows the performance of different methods in terms of accuracy, model size, and inference time. Our method achieves a good balance between small model sizes and excellent results while having a comparable inference time that is better than SOTA baselines. Although its response time is worse than traditional handwriting authentication methods, its accuracy is much higher. Compared with self-supervised learning and advanced handwriting authentication baseline methods, it also matches or exceeds SOTA methods both in model size and inference time. In summary, our method outperforms SOTA methods in all aspects, i.e., handwriting authentication performance, model size, and inference time.

\subsection{Robustness Analysis}
\label{sec:6.3}
In this subsection, we analyze the robustness of SherlockNet. We evaluate the anti-interference ability of SherlockNet from two perspectives: data and features.

\subsubsection{\textbf{Robust Analysis for Data}}
Freeform handwriting authentication in reality faces problems of forgery and defects, such as immersion, stains, paper damage, etc. To validate the robustness of SherlockNet, we conduct comparative experiments based on damaged data and falsified data, adopting the EN-HA dataset specially constructed for real-world applications. We adjust the ratio of damaged and falsified data in EN-HA and record the average results. We select ten baselines with better performance in Table \ref{tab:comparison} for comparison. 

Table \ref{tab:robustness} shows the performance of different methods under non-ideal data. The results show that the performance of SherlockNet does not change much in data containing a lot of noise, proving its advantages in robustness. Specifically, when the proportion of falsified data increases by 10\%, the performance of SherlockNet is almost unchanged, while the performance of other methods all decreases. When the proportion of falsified data increases to 30\%, the advantage of SherlockNet expands to more than 15\%. In the case of damaged data, when the noise ratio reaches 50\%, the performance of SherlockNet only drops by 1\%, exceeding other methods. These findings validate the robustness of SherlockNet and its advantages in practical applications, especially in historical digital archives where the damage of manuscripts due to improper storage and transportation is typically more than 30\%.

\subsubsection{\textbf{Robust Analysis for Feature}}
Considering that the augmentation of SSL may affect the features, e.g., affine transformation may cause deformation of the notes, and color transformation may make the model consider color information when classifying. Therefore, we choose five augmentation methods that affect the handwriting data \cite{zoph2020learning}, including color transformation, noise addition, affine transformation, cropping, and brightness adjustment, denoted as $A_1\sim A_5$. We record the performance of SherlockNet and the self-supervised baselines when using these augmentation methods on three conditions, where we let the differences between the augmented image and the pre-augmented image be $\le 20\%$ (Group 1), $20\sim 40\%$ (Group 2), and $\ge 40\%$ (Group 3) respectively.

Figure \ref{fig:5.4} shows the experimental results. The results show that SherlockNet can maintain stable performance under various augmentation methods, while other SSL baselines have different degrees of performance degradation. We think this is because (i) color transformation and brightness adjustment may cause the model to mistakenly use color and brightness as discriminative information, (ii) noise addition and cropping may increase the difficulty of self-supervised baseline methods to cope with noise where cropping changes the proportion of interference, (iii) affine transformation changes the size and shape of the font, while SSL baselines rely too much on it and ignore the handwriting style. Therefore, the results demonstrate that SherlockNet can effectively extract the most critical features and perform robust handwriting authentication.

%% ---Table 3 Methods used for Ablation Study-----
\begin{table*}
  \caption{Ablation study of SherlockNet on EN-HA. The ``$\checkmark$" indicates that the corresponding module is activated in this round of testing. The ``D-N\%" and ``F-N\%" after ``Accuracy(\%)" represent the area ratio of noise and the ratio of falsified samples increased in the EN-HA participating in the experiment, respectively.}
  \label{tab:ablation}
  \begin{tabular}{l|l|ccccccc}
    \toprule
    \multirow{4}{*}{Extractors} & ViT     &$\checkmark$&$\checkmark$&$\checkmark$&$\checkmark$&  &  &  \\
    & Conv4 &  &  &  &  & $\checkmark$ &  &  \\
    & Resnet-50 &  &  &  &  &  &$\checkmark$ &  \\
    & Swin &  &  &  &  &  &  &$\checkmark$ \\
    \cline{1-2}   
    \multicolumn{2}{l|}{Energy-oriented Operator} &  &$\checkmark$ & &$\checkmark$&$\checkmark$&$\checkmark$&$\checkmark$\\   
    \multicolumn{2}{l|}{Two-branch Paradigm} &  &  & $\checkmark$ &$\checkmark$ &$\checkmark$&$\checkmark$&$\checkmark$ \\
    \midrule    
    \multicolumn{2}{l|}{\textbf{Accuracy(\%)}} & 76.42 $\pm$ 0.24 & 79.95 $\pm$ 0.27 & 81.47 $\pm$ 0.14 & 82.78 $\pm$ 0.35 & 80.57 $\pm$ 0.33 & 80.90 $\pm$ 0.63 & 83.46 $\pm$ 0.10  \\
    \multicolumn{2}{l|}{\textbf{Accuracy(\%)-D-10\%}} & 75.54 $\pm$ 0.18 & 79.46 $\pm$ 0.19 & 81.09 $\pm$ 0.13 & 82.89 $\pm$ 0.24 & 80.54 $\pm$ 0.15 & 82.65 $\pm$ 0.14 & 83.89 $\pm$ 0.10 \\
    \multicolumn{2}{l|}{\textbf{Accuracy(\%)-D-30\%}} & 73.07 $\pm$ 0.09 & 77.32 $\pm$ 0.17 & 79.45 $\pm$ 0.17 & 81.92 $\pm$ 0.15 & 79.84 $\pm$ 0.28 & 80.12 $\pm$ 0.17 & 81.89 $\pm$ 0.38 \\
    \multicolumn{2}{l|}{\textbf{Accuracy(\%)-D-50\%}} & 70.57 $\pm$ 0.25 & 76.28 $\pm$ 0.18 & 77.51 $\pm$ 0.16 & 81.07 $\pm$ 0.03 & 79.28 $\pm$ 0.09 & 80.17 $\pm$ 0.77 & 83.25 $\pm$ 0.10 \\
    \multicolumn{2}{l|}{\textbf{Accuracy(\%)-F-10\%}} & 73.78 $\pm$ 0.28 & 73.02 $\pm$ 0.09 & 81.56 $\pm$ 0.18  & 82.28 $\pm$ 0.32 & 81.21 $\pm$ 0.15 & 81.35 $\pm$ 0.13 & 83.18 $\pm$ 0.24 \\
    \multicolumn{2}{l|}{\textbf{Accuracy(\%)-F-20\%}} & 70.94 $\pm$ 0.17 & 70.24 $\pm$ 0.15 & 80.84 $\pm$ 0.15 & 81.37 $\pm$ 0.04 & 79.54 $\pm$ 0.51 & 81.86 $\pm$ 0.15 & 80.50 $\pm$ 0.09 \\
    \multicolumn{2}{l|}{\textbf{Accuracy(\%)-F-30\%}} & 65.23 $\pm$ 0.21 & 66.24 $\pm$ 0.05 & 78.16 $\pm$ 0.37 & 79.45 $\pm$ 0.16 & 76.27 $\pm$ 0.09 & 77.87 $\pm$ 0.17 & 78.85 $\pm$ 0.16 \\
    \bottomrule
  \end{tabular}
\end{table*}

\subsection{Ablation Study}
\label{sec:6.4}

In this subsection, we conduct ablation studies to explore the effect of different modules of SherlockNet and the selection of hyperparameters $\lambda_1$ and $\lambda_2$ in Eq.\ref{equ:loss}, respectively.

\textbf{The effect of different modules.} We conduct ablation studies to analyze the impact of the three components of SherlockNet, including the feature extractor, energy-oriented operator, and two-branch momentum-based adaptive contrastive learning paradigm (two-branch paradigm). Specifically, for the feature extractor, we try four types of feature extractors mentioned in Subsection \ref{sec:6.1}. For the energy-oriented operator, we directly disabled this module. For the two-branch learning paradigm, we replace our method with the traditional method but with the same extractor and the energy-oriented operator \cite{chen2020simple}. Moreover, we simulate non-ideal scenarios consistent with Subsection \ref{sec:6.3} for further exploration.
Table \ref{tab:ablation} shows the results. We can observe that: (i) for the feature extractors, the hierarchical Transformers structure helps to eliminate redundant patches, and Swin achieves higher detection accuracy. (ii) the energy-oriented operator and the two-branch paradigm can effectively improve the model performance, achieving about 3\% and 7\% improvement on EN-HA respectively. This indicates that our design is foresighted.

\textbf{The selection of $\lambda_1$ and $\lambda_2$.} The hyperparameters $\lambda _1$ and $\lambda _2$ in Eq.\ref{equ:loss} are the importance of two different branch losses, i.e., loss of the contrastive learning branch $\mathcal{L}_{cl}$ and the loss of the momentum branch $\mathcal{L}_{mo}$, in the overall optimization loss $\mathcal{L}$. It is worth noting that the sum of $\lambda _1$ and $\lambda _2$ is 1, that is, $\lambda _2=1-\lambda _1$. We evaluate the performance (accuracy(\%)) of SherlockNet with different $\lambda_1$ and $\lambda_2$ on EN-HA, following the same implementation discussed in Subsection \ref{sec:6.2}. 
The results in Figure \ref{fig:ablation_2} show that the performance is optimal when $\lambda_1=0.6$ and $\lambda_2=0.4$, which is also the hyperparameter settings of the proposed SherlockNet.

\subsection{Visualization Analyses}
\label{sec:6.5}

\subsubsection{\textbf{Feature Visualization}}
\begin{figure}
    \centering
    \begin{subfigure}{0.15\textwidth}
        \includegraphics[width=\textwidth]{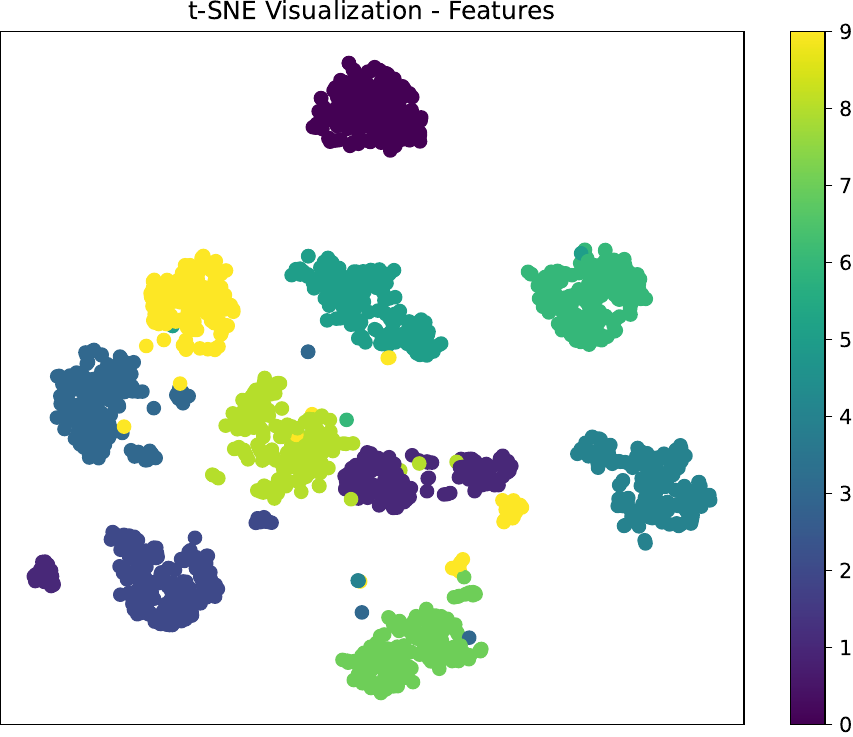}
        \caption{}
        \label{fig:e-sub1}
    \end{subfigure}
    \hfill
    \begin{subfigure}{0.15\textwidth}
        \includegraphics[width=\textwidth]{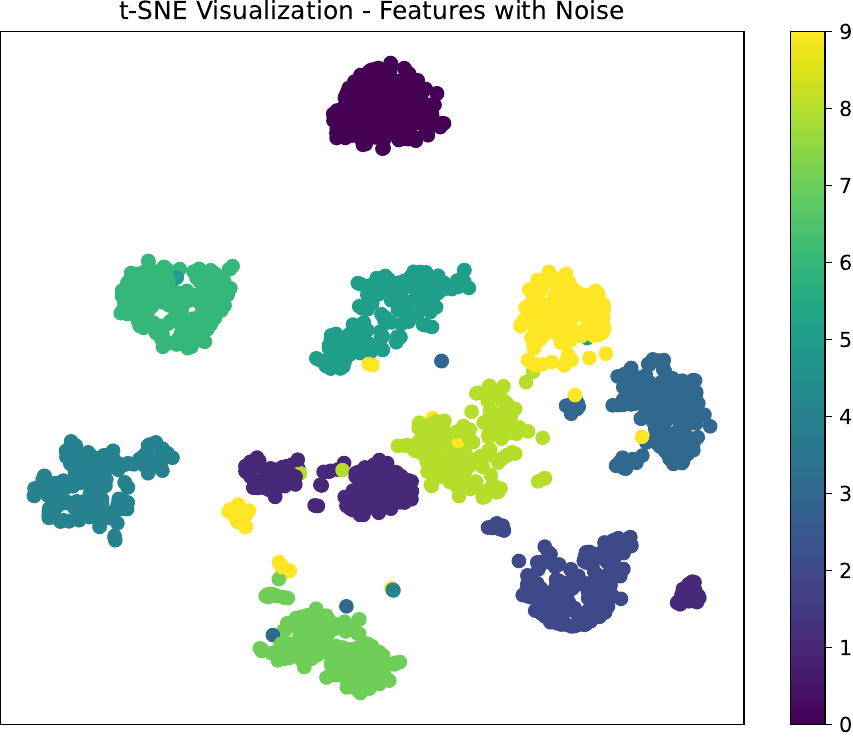}
        \caption{}
        \label{fig:e-sub2}
    \end{subfigure}
    \hfill
    \begin{subfigure}{0.15\textwidth}
        \includegraphics[width=\textwidth]{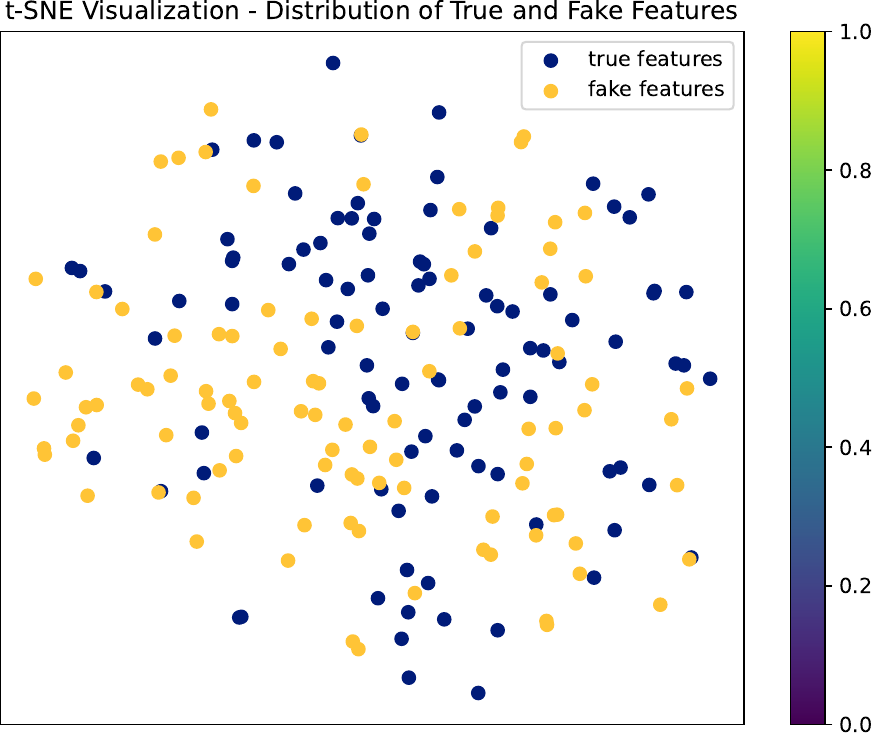}
        \caption{}
        \label{fig:e-sub3}
    \end{subfigure}
    \caption{Feature visualization. (a) shows the t-SNE visualization of handwriting features, (b) shows the t-SNE visualization of handwriting features with noise, and (c) shows the t-SNE visualization of the feature distribution in true and fake data.}
    \label{fig:E_1}
\end{figure}

One challenge of freeform handwriting authentication is the complexity of features. To explore the learning process in depth, we used t-SNE to visualize the feature representations of EN-HA. We visualize three scenarios: 100\% ideal data, 70\% ideal data + 30\% forged data, and data with 30\% noise, respectively. Figure \ref{fig:E_1} shows the feature distribution of the three scenarios. From the results, we can observe that: (i) handwriting data have diverse features, and some features are difficult to distinguish; (ii) in redundant data, noise shows intra-class aggregation, but there is no significant difference from real features, which may cause some models to mistake noise as real features and learn from them; and (iii) the feature distribution of forged data is different from that of real data, but the difference is small and hard to be detected. Despite this, SherlockNet still achieves robust discrimination as shown in Subsection \ref{sec:6.2}, demonstrating its robustness.

\subsection{Discussion and Future Work}
\label{sec:6.6}
SherlockNet has shown superior results in freeform handwriting authentication even with various interference. However, despite the above breakthroughs, freeform handwriting authentication remains an open area for research, which we aim to explore further in the future. Firstly, changes in handwriting may occur due to various reasons, i.e., mood, mental state, age, etc. Additionally, different writing instruments can lead to different handwriting styles. Therefore, we plan to introduce the temporal dimension into freeform handwriting authentication, rather than limiting it to the identification of individual style fixations. Moreover, supervised methods for freeform handwriting authentication are task-specific due to the different writing properties of different languages, e.g., the cursive form of Arabic. SherlockNet is suitable for such complex and challenging changes, but the lack of evaluation standards and more benchmark datasets hinders its development. We believe this work can provide a solid foundation for freeform handwriting authentication, and we will further expand our research in the future.

%% -------------------Conclusion--------------------

\section{Conclusion}
\label{Conclusion}
In this paper, we propose SherlockNet, a novel energy-oriented two-branch contrastive self-supervised learning framework for robust and fast freeform handwriting authentication. It addresses the three key challenges in freeform handwriting authentication, i.e., (i) severe damage, (ii) complex high-dimensional features, and (iii) lack of supervision. SherlockNet consists of four stages, i.e., pre-processing, generalized pre-training, personalized fine-tuning, and practical application. Specifically, in the pre-processing stage, we develop a plug-and-play energy-oriented operator to calculate the energy distribution in each handwriting manuscript, eliminating the impact of data corruption and forgery such as scratches and stains. In the pre-training stage, we propose a two-branch momentum-based adaptive contrastive learning framework to learn general representations from energy distributions, enabling swift extraction of complex high-dimensional features while also identifying spatial correlations. Moving into the personalized fine-tuning and practical application stages, we develop user-friendly interfaces that allow individuals to easily deploy SherlockNet in various real-world applications. This deployment merely necessitates a few samples with a few steps, facilitating effortless freeform handwriting authentication. 
Moreover, to simulate real-world scenarios, we construct a new dataset called EN-HA, which contains damaged and forged data.
Extensive experiments demonstrate that SherlockNet outperforms existing baselines, highlighting its effectiveness and robustness for freeform handwriting authentication.

\bibliographystyle{IEEEtran}
\bibliography{TMM}

\begin{IEEEbiography}[{\includegraphics[width=1in,height=1.25in,clip,keepaspectratio]{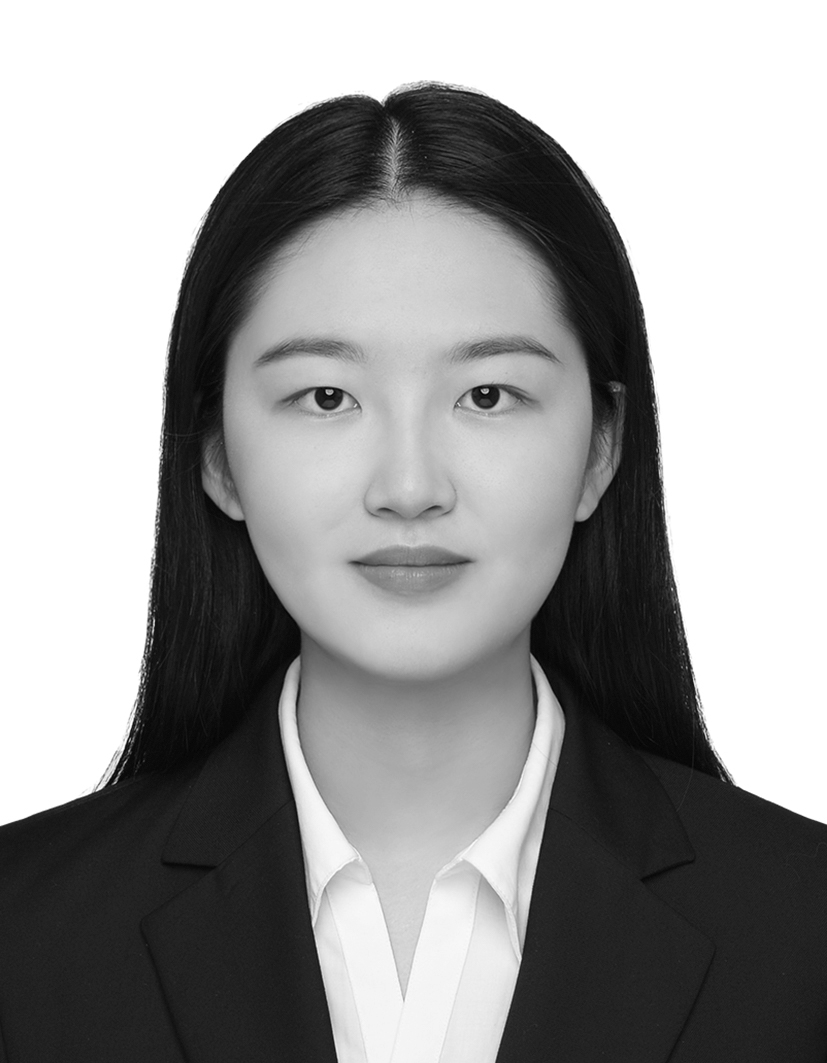}}]{Jingyao Wang}
	received the B.S. degree in Robotics Engineering from Beijing University of Technology in 2018. She is currently a postgraduate student at the University of Chinese Academy of Sciences, Beijing, China. She is also with the Science \& Technology on Integrated Information System Laboratory, Institute of Software Chinese Academy of Sciences, Beijing, China. Her research interests include transfer learning, meta-learning, robot learning, and machine learning.
\end{IEEEbiography}
% \vspace{-0.2in}

\begin{IEEEbiography}[{\includegraphics[width=1in,height=1.25in,clip,keepaspectratio]{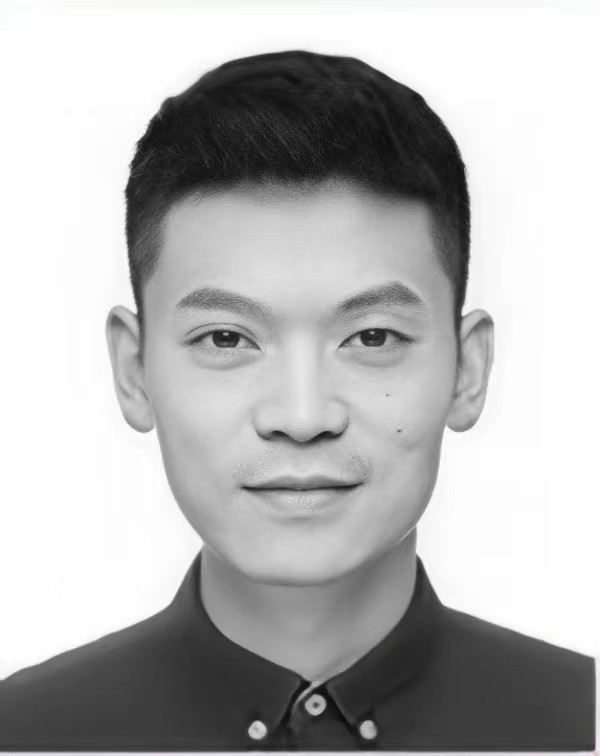}}]{Luntian Mou}
	(M’12-SM’22) received a Ph.D. degree in computer science from the University of Chinese Academy of Sciences, Beijing, China, in 2012. He accomplished as a Postdoctoral Researcher with the Institute of Digital Media, Peking University, Beijing, China, in 2014. From 2019 to 2020, he served as a visiting scholar at the Donald Bren School of Information and Computer Sciences, University of California, Irvine, CA, USA. He is currently an Associate Professor with the Beijing Key Laboratory of Multimedia and Intelligent Software Technology, the Faculty of Information Technology, Beijing University of Technology, Beijing, China. He is the founding chair of IEEE Workshop on Artificial Intelligence for Art Creation (AIART). His research interests include artificial intelligence, machine learning, brain-like computing, multimedia computing, affective computing, and neuroscience.
\end{IEEEbiography}
% \vspace{-0.2in}

\begin{IEEEbiography}[{\includegraphics[width=1in,height=1.25in,clip,keepaspectratio]{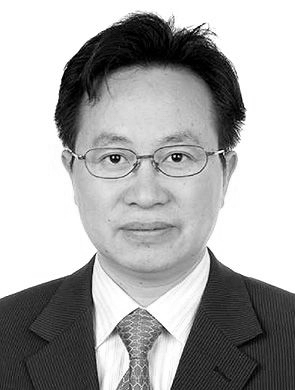}}]{Changwen Zheng}
	received a B.S. degree in
mathematics from Huazhong Normal
University in 1992 and the Ph.D. degree in computer science and technology from Huazhong University of Science and Technology in 2003, respectively. He is is currently a Professor with the Institute of Software Chinese Academy of Sciences. His current research interests include route planning, evolutionary computation, and neural networks.
\end{IEEEbiography}
% \vspace{-0.2in}

 \begin{IEEEbiography}[{\includegraphics[width=1in,height=1.25in,clip,keepaspectratio]{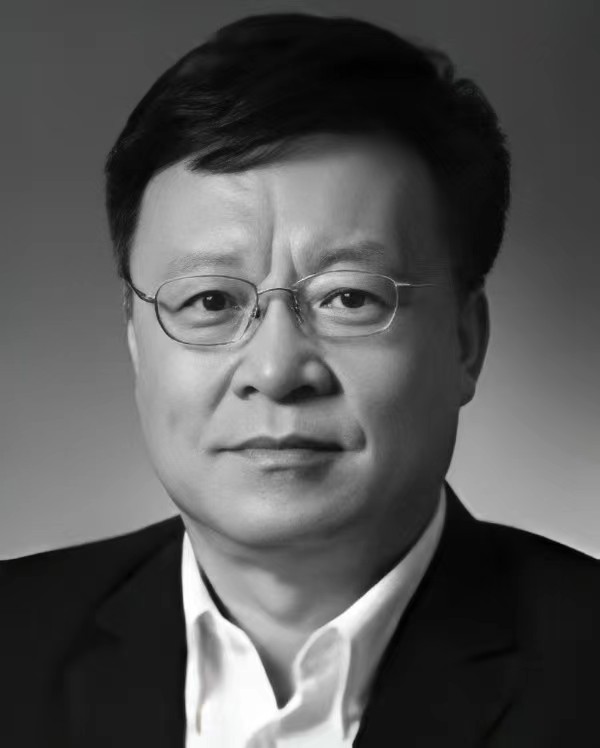}}]{Wen Gao}
 	(M’92-SM’05-F’09) received the Ph.D. degree in electronic engineering from The University of T okyo, Japan, in 1991. He is currently a Professor of computer science with Peking University, Beijing, China. He has served on the Editorial Boards for several journals, such as the IEEE T ransactions on Circuits and Systems for Video T echnology, the IEEE T ransactions on Multimedia, the IEEE T ransactions on Autonomous Mental Development, the Eurasip Journal of Image Communications, and the Journal of Visual Communication and Image Representation. He chaired a number of prestigious international conferences on multimedia and video signal processing, such as the IEEE ISCAS, ICME, and the ACM Multimedia, and also served on the advisory and technical committees of numerous professional organizations.
 \end{IEEEbiography}

\vfill

\end{document}